\documentclass{article}

\usepackage{adjustbox}
\usepackage{tabularx}

\usepackage[final]{neurips_2024}
\usepackage[utf8]{inputenc} 
\usepackage[T1]{fontenc}    
\usepackage[dvipsnames]{xcolor}   
\usepackage[colorlinks=true,
            linkcolor=Blue,
            urlcolor=Blue,
            citecolor=Blue,
            anchorcolor=Blue]{hyperref}       %
\usepackage{xurl}            
\usepackage{booktabs}       
\usepackage{amsfonts}       
\usepackage{nicefrac}       
\usepackage{microtype}      
\usepackage{amsmath} 
\usepackage{amssymb}      
\usepackage{tikz}
\usepackage{bm}
\usetikzlibrary{automata, positioning, arrows}
\usepackage{environ}
\usepackage{subcaption}
\usepackage{placeins}
\usepackage{multirow}
\usepackage{colortbl}
\usepackage{makecell}
\usepackage{float}
\usepackage{wrapfig}
\usepackage{enumitem}
\setitemize{noitemsep,topsep=0pt,parsep=0pt,partopsep=0pt}
\title{Exploring Consistency in Graph Representations: from Graph Kernels to Graph Neural Networks}
\newtheorem{theorem}{Theorem}[section]

\newtheorem{lemma}[theorem]{Lemma}

\newtheorem{definition}[theorem]{Definition}
\newtheorem{proof}{Proof}
\newcommand{\empcell}[1]{\cellcolor[RGB]{240,255,255} #1}  
\author{
    Xuyuan Liu$^\ast$ \hspace{2em} Yinghao Cai$^\ast$ \hspace{2em} Qihui Yang$^{\dag}$ \hspace{2em} Yujun Yan$^\ast$\\
     Dartmouth College$^\ast$\\
    \texttt{\{xuyuan.liu.gr, yinghao.cai, yujun.yan\}@dartmouth.edu}\\
    University of California San Diego$^{\dag}$\\
    \texttt{qiy009@ucsd.edu}
}
\begin{document}
\maketitle
\vspace{-0.45cm}
\begin{abstract}
 Graph Neural Networks (GNNs) have emerged as a dominant approach in graph representation learning, yet they often struggle to capture consistent similarity relationships among graphs. 
 While graph kernel methods such as the Weisfeiler-Lehman subtree (WL-subtree) and Weisfeiler-Lehman optimal assignment (WLOA) kernels are effective in capturing similarity relationships, they rely heavily on predefined kernels and lack sufficient non-linearity for more complex data patterns. Our work aims to bridge the gap between neural network methods and kernel approaches by enabling GNNs to consistently capture relational structures in their learned representations. Given the analogy between the message-passing process of GNNs and WL algorithms, we thoroughly compare and analyze the properties of WL-subtree and WLOA kernels. We find that the similarities captured by WLOA at different iterations are asymptotically consistent, ensuring that similar graphs remain similar in subsequent iterations, thereby leading to superior performance over the WL-subtree kernel. Inspired by these findings, we conjecture that the consistency in the similarities of graph representations across GNN layers is crucial in capturing relational structures and enhancing graph classification performance. Thus, we propose a loss to enforce the similarity of graph representations to be consistent across different layers. Our empirical analysis verifies our conjecture and shows that our proposed consistency loss can significantly enhance graph classification performance across several GNN backbones on various datasets. 
\let\thefootnote\relax\footnotetext{Code: \url{https://github.com/GraphmindDartmouth/Graph-consistency}}
\end{abstract}
\vspace{-0.1cm}
\section{Introduction}
\begin{figure}[h]
    \centering
    \includegraphics[width=0.62\textwidth]{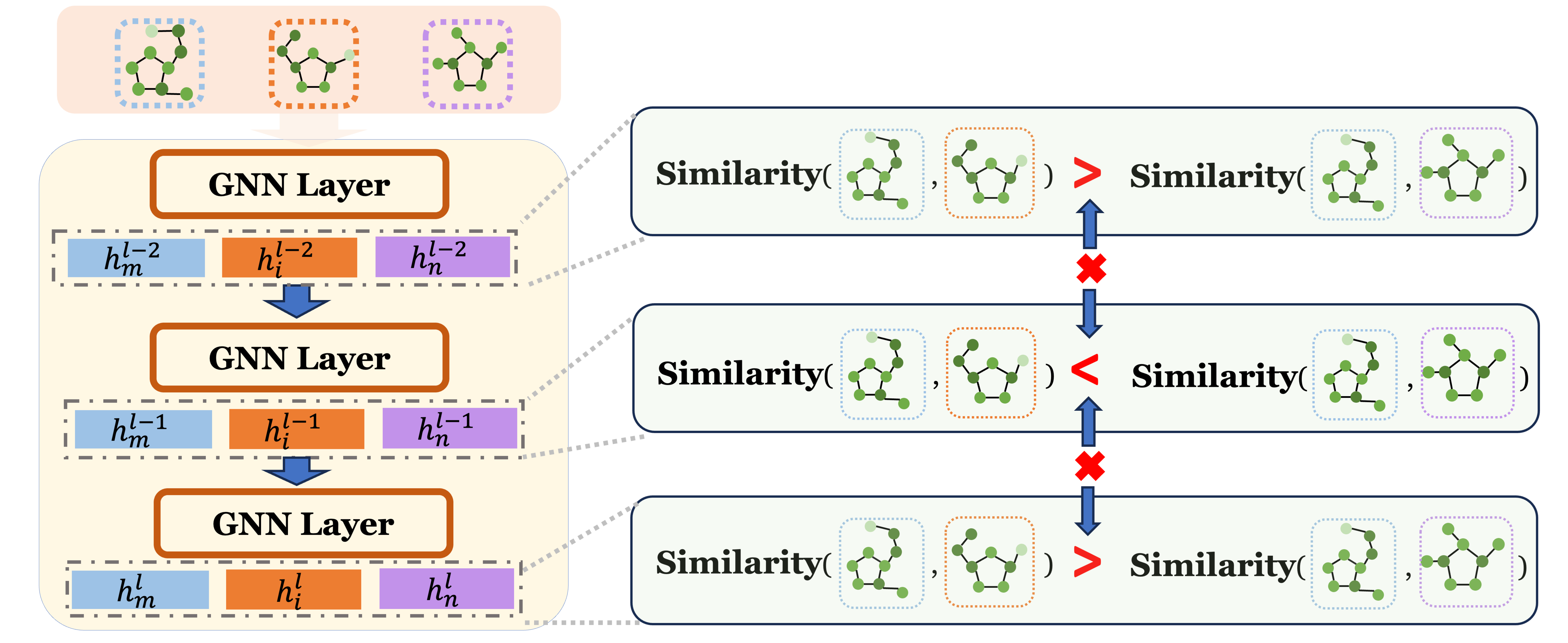}
    \caption{\small Cosine similarity of three molecules from the NCI1 dataset, evaluated using graph representations from three consecutive GIN layers. Common GNN models fail to  preserve relational structures across the layers.} 
    \vspace{-0.65cm}
    \label{fig:Intro_simialrity}
\end{figure}

\vspace{-0.1cm}
Graph classification tasks are extensively applied across multiple domains, including chemistry~\citep{DBLP:conf/iclr/LiuWLLGT22,DBLP:conf/icml/XuPDEL23}, bioinformatics~\citep{yan2019groupinn, li2023interpretable, li2023size}, and social network analysis~\citep{DBLP:conf/kdd/YingHCEHL18, wangevolunet}. Graph neural networks (GNNs)~\citep{DBLP:conf/iclr/KipfW17, DBLP:conf/iclr/XuHLJ19, DBLP:conf/iclr/VelickovicCCRLB18, huangenhancing} have emerged as the predominant approach for performing graph classification, owing to their ability to extract rich representations from various types of graph data. A typical GNN employs the message-passing mechanism~\citep{DBLP:conf/icml/GilmerSRVD17}, where node features are propagated and aggregated across connected nodes. This process effectively captures local tree structures, enabling the differentiation between various graphs. However, GNNs often struggle to  preserve relational structures among graphs, resulting in inconsistent relative similarities across the layers.
As shown in Figure~\ref{fig:Intro_simialrity}, graphs with higher relative similarity in one layer may exhibit reduced similarity in the subsequent layer. This phenomenon arises from the limitations of cross-entropy loss, which fails to preserve relational structures, as it forces graphs within the same class into identical representations.

Graph kernel methods, on the other hand, are designed to capture similarities between graphs and utilize these similarities for classification tasks. For instance, subgraph-pattern approaches~\citep{DBLP:journals/jmlr/ShervashidzeVPMB09,DBLP:journals/ans/KriegeJM20} compare graphs by counting the occurrences of fixed-size subgraph motifs. Other methods compare sequences of vertices or edges encountered during graph traversals~\citep{DBLP:conf/icdm/BorgwardtK05,DBLP:conf/icml/KashimaTI03,DBLP:conf/nips/0007WX0N18}. Among all graph kernels, two notable ones are the Weisfeiler-Lehman subtree (WL-subtree) kernel~\citep{DBLP:journals/jmlr/ShervashidzeSLMB11} and the Weisfeiler-Lehman optimal assignment (WLOA) kernel~\citep{DBLP:conf/nips/KriegeGW16}. They are found to have comparable performance to simple GNNs~\citep{DBLP:journals/jair/NikolentzosSV21}.
The WL-subtree kernel iteratively relabels graphs using the Weisfeiler-Lehman algorithm~\citep{weisfeiler1968reduction} and constructs a kernel based on the number of occurrences of each label. The WLOA kernel uses the same relabeling scheme but computes a matching between substructures to reveal structural correspondences between the graphs. 

While effective in capturing relative graph similarity, kernel methods rely on predefined kernels and exhibit insufficient non-linearities, limiting their ability to capture complex patterns in high-dimensional data. Additionally, kernel methods are computationally costly, making them unsuitable for handling large datasets and consequently limiting their overall applicability.

In this work, we aim to bridge the gap between kernel methods and GNN models.  Given the iterative nature of GNNs, we
study a class of kernels which are induced from graph representations obtained through an iterative process and name them iterative graph kernels (IGK). Within this framework, we define the \textbf{consistency} property, which ensures that similar graphs remain similar in subsequent iterations. Our analysis demonstrates that kernels with this property ensure better generalization ability, leading to improved classification performance.Furthermore, we find that this property sheds light on why the WLOA kernel outperforms the WL-subtree kernel. The WLOA kernel asymptotically demonstrates consistency as the iteration goes to infinity, whereas the WL-subtree kernel does not exhibit this behavior. Inspired by these findings and the analogy between message-passing GNNs and the WL-subtree kernel~\citep{DBLP:conf/iclr/XuHLJ19},
we hypothesize that this principle is also applicable to GNNs. To explore this, we introduce a novel loss function designed to align the ranking of graph similarities across GNN layers. The aim is to ensure that the relational structures of the graphs are preserved and consistently reflected throughout the representation space of these layers. We validate this hypothesis by applying our loss function to different GNN backbones across various graph datasets. Extensive experiments demonstrate that our proposed model-agnostic consistency loss improves graph classification performance comprehensively.

In summary, the main contributions of this work are as follows:
\begin{itemize}
    \item \textbf{Novel perspective:} 
    We present a novel perspective on understanding the graph classification performance of GNNs by analyzing the similarity relationships captured by different layers.
    \item \textbf{New insights:} 
    {We are the first to introduce and formalize the consistency principle within both kernel-based and GNN methods for graph classification tasks. Additionally, we provide theoretical proofs explaining how this principle enhances the performance.}
    \item \textbf{Simple yet effective method:}  {Empirical results demonstrate that the proposed consistency loss universally enhances performance across a wide range of base models and datasets.}
\end{itemize}
\vspace{-0.05cm}
\vspace{-0.20cm}
\section{Preliminaries}
\vspace{-0.20cm}

In this section, we begin by introducing the notations and definitions used throughout the paper. Next, we provide an introduction to the fundamentals of Weisfeiler-Lehman isomorphism test, GNNs and graph kernels.
\vspace{-3pt}
\subsection{Notations and Definitions}
\vspace{-3pt}
Let $\mathcal{G}$($\mathcal{V}$, $\mathcal{E}$, $\mathbf{X}$) be an undirected and unweighted graph with $N$ nodes, where $\mathcal{V}$ denotes the node set, $\mathcal{E}$ denotes the edge set, and $\mathbf{X}$ denotes the feature matrix, where each row represents the features of a corresponding node. The neighborhood of a node $v$ is defined as the set of all nodes that connected to $v$: $\mathcal{N}(v)=\{u|(v, u) \in \mathcal{E}\}.$ In a graph dataset, each graph $\mathcal{G}_i$ is associated with a label $\mathcal{Y}_i$, which is sampled from a label set ${\mathcal{L}}$. In this paper, we focus on the graph classification task, where a model $\phi$ is trained to map each graph to its label. 
\vspace{-3pt}
\subsection{Weisfeiler-Lehman Isomorphism Test}
\vspace{-3pt}
We first introduce the Weisfeiler-Lehman isomorphism test~\citep{weisfeiler1968reduction}, which can be used to distinguish different graphs and is closely related to the message-passing process of GNNs~\citep{DBLP:conf/iclr/XuHLJ19}. The WL algorithm operates by iteratively relabeling the colors of vertices in the graph. Initially, all vertices are assigned the same color $C_0$. In iteration $i$, the color of a vertex $v$ is updated based on its current color $C_{v, i-1}$ and the colors of its neighbors $\left\{C_{u \in \mathcal{N}({v}), i-1}\right\}$. The update is given as follows:
$$
C_{{v}, i}=f_c^{i}\left(\left\{C_{{v}, i-1},\left\{C_{u \in \mathcal{N}({v}), i-1}\right\}\right\}\right)
$$
where $f_c^{i}$ is an injective coloring function that  maps the multisets to different colors at iteration $i$.This process continues for a predefined number of iterations or until the coloring stabilizes (i.e., the colors no longer change).
\vspace{-3pt}
\subsection{Graph Neural Network}
\vspace{-3pt}
Most GNNs adopt the message-passing 
framework \citep{DBLP:conf/icml/GilmerSRVD17}, 
which can be viewed as a derivative of the Weisfeiler-Lehman coloring mechanism. Specifically, let $\boldsymbol{h}_v^{(k-1)}$ represent the feature vector of node $v$ at the $(k-1)$-th iteration. A GNN computes the new feature for $v$ by aggregating the representations of itself and its neighboring nodes $u \in \mathcal{N}(v)$ as follows:
$$
\boldsymbol{h}_v^{(k)}=\texttt{UPDATE}^{(k)}\left(\boldsymbol{h}_v^{(k-1)}, \boldsymbol{m}_v^{(k)}\right) \text {, where } \boldsymbol{m}_v^{(k)}=\texttt{AGGR}^{(k)}\left(\left\{\boldsymbol{h}_u^{(k-1)}: u \in \mathcal{N}(v)\right\}\right)
$$
The initial node representations $\boldsymbol{h}_v^{(0)}$ are set to the raw node features $\mathbf{X}_v$. At the $k$-th iteration, the aggregation function $\texttt{AGGR}^{(k)}(\cdot)$ computes the messages $\boldsymbol{m}_v^{(k)}$ received from neighboring nodes.
Subsequently, the update function $\texttt{UPDATE}^{(k)}(\cdot)$ computes a new representation for each node by integrating the neighborhood messages $\boldsymbol{m}_v^{(k)}$ with its previous embedding $\boldsymbol{h}_v^{(k-1)}$. After $T$ iterations, the final node representations are combined into a graph representation using a readout function:
$$
\boldsymbol{h}_G=\texttt{READOUT}\left(\left\{\boldsymbol{h}_v^{(T)} \mid v \in \mathcal{V}\right\}\right) .
$$
The readout function, essentially a set function learned by the neural network, commonly employs \texttt{AVERAGE} or \texttt{MAXPOOL}.
\vspace{-3pt}
\subsection{Graph Kernel}\label{subsec:gkernel}
\vspace{-3pt}
A kernel is a function used to measure the similarity between pairs of objects. For a non-empty set $\chi$ and a function $\mathbb{K}: \chi \times \chi \rightarrow \mathbb{R}$, the function $\mathbb{K}$ qualifies as a kernel on $\chi$ if there exists a Hilbert space $\mathcal{H}_k$ and a feature map function $\phi: \chi \rightarrow \mathcal{H}_k$, such that $\mathbb{K}(x, y)=\langle\phi(x), \phi(y)\rangle$ for any $x, y \in \chi$, where $\langle\cdot, \cdot\rangle$ denotes the inner product in $\mathcal{H}_k$. Notably, such a feature map exists if and only if $\mathbb{K}$ is a positive semi-definite function.
Let $\mathbb{K}$ be a kernel defined on $\chi$, and let
$S=\left\{x_1, \ldots, x_n\right\}$ be a finite set of $n$ samples on $\chi$. 
The Gram matrix for $S$ is defined as $\mathbf{G} \in \mathbb{R}^{n \times n}$, 
with each element $\mathbf{G}_{ij} = \mathbb{K}(x_i, x_j)$ representing the kernel value between the $i$-th and $j$-th data points in $S$. The Gram matrix is always positive semi-definite.

Graph kernel methods apply the kernel approaches to graph data, typically defined using the $\mathcal{R}$-convolution framework \citep{haussler1999convolution,DBLP:conf/log/BauseK22}. Consider two graphs, $\mathcal{G}$ and $\mathcal{G}^{\prime}$. The key idea is to decompose the graphs into substructures using a predefined feature map function $\phi$, and then compute the kernel value by taking the inner product in the feature space: $\mathbb{K}(\mathcal{G}, \mathcal{G}^{\prime}) = \langle \phi(\mathcal{G}), \phi(\mathcal{G}^{\prime}) \rangle$, based on these substructures.
Weisfeiler-Lehman (WL) graph kernels stand out as one of the most widely used approaches. These methods employ the Weisfeiler-Lehman coloring scheme to iteratively encode graphs, calculating kernel values based on these colorings. The final kernel values are derived through the aggregation of intermediate results. Next, we introduce the WL-subtree kernel and the WLOA kernel in more detail. Let $f^{i}$ be the coloring function at the $i$-th iteration, mapping the colored graph from the previous iteration to a new colored graph. Define $\psi^{i}$ as the function that captures the cumulative coloring effect up to the  $i$-th iteration:  $\psi^{i}=f^{i} \circ \cdots \circ f^{1}$.
Specifically, the WL-subtree kernel computes the kernel value by directly using the label histogram as a feature to compute the dot product, which is expressed as: 
\begin{equation*}
    \mathbb{K}^{(h)}_{wl\_subtree}\left(\mathcal{G}, \mathcal{G}^{\prime}\right)=\sum_{i=1}^h \left\langle\phi(\psi^{i}(\mathcal{G})), \phi(\psi^{i}\left(\mathcal{G}^{\prime})\right)\right\rangle
    \label{equa:wlsubtree}
\end{equation*}
The WLOA kernel applies a histogram intersection kernel to match substructures between different graphs,  which is formulated as:
\begin{equation*}
    \mathbb{K}^{(h)}_{WLOA}\left(\mathcal{G}, \mathcal{G}^{\prime}\right)=\sum_{i=1}^h \texttt{histmin} \left\{\phi(\psi^{i}(\mathcal{G})), \phi(\psi^{i}(\mathcal{G}^{\prime}))\right\} \cdot \omega(i)
    \label{equation:wloa}
\end{equation*}
where $\omega(i)$ is a nonnegative, monotonically non-decreasing weight function, and $\omega(i)=1$ is commonly used in practice~\citep{DBLP:conf/nips/KriegeGW16,DBLP:journals/jmlr/SiglidisNLGSV20}. {The operator $\texttt{histmin}$ denotes the histogram intersection kernel. It is computed by comparing and summing the smallest matching elements between two sets. For example, consider two sets $S_1:\{ \mathfrak{a},\mathfrak{a},\mathfrak{b},\mathfrak{b},\mathfrak{c} \}$ and $S_2:\{ \mathfrak{a},\mathfrak{b},\mathfrak{b},\mathfrak{c},\mathfrak{c} \}$. The $\texttt{histmin}(S_1,S_2)$ is calculated by taking the minimum frequency of each distinct element across the two sets, yielding $\min(2,1) + \min(2,2) + \min(1,2) = 4$.}
To ensure that the kernel values fall in the range of 0 to 1, normalization is often applied. The normalized kernel value $\tilde{\mathbb{K}}^{(h)}$ is then expressed as follows:
\begin{equation*}
    \tilde{\mathbb{K}}^{(h)}(\mathcal{G}, \mathcal{G}^{\prime })=\frac{\mathbb{K}^{(h)}(\mathcal{G}, \mathcal{G}^{\prime })}{\sqrt{\mathbb{K}^{(h)}(\mathcal{G}, \mathcal{G})}{\sqrt{\mathbb{K}^{(h)}(\mathcal{G}^{\prime }, \mathcal{G}^{\prime })}}}
    \label{equ:normalized_wl}
\end{equation*}

\vspace{-5pt}
\section{Consistency Principles}
\vspace{-0.1cm}
To encode relational structures in GNN learning, we first examine how similarities are represented in graph kernels. In this section, we start by defining a class of graph kernels, i.e., the iterative graph kernels, which encompasses many widely used kernels. Then, we delve into a key property known as the consistency property, which may play an important role in enhancing classification performance. We support this assertion through theoretical analysis, elucidating how different kernels adhere to or deviate from this property, thereby explaining their performance differences.
\subsection{Iterative Graph Kernels}\label{section:definition_kernel}
\vspace{-0.1cm}
In this paper, we are interested in a set of kernels defined as follows:
\begin{definition}
Given a colored graph set $\chi$, a feature map function $\phi : \chi \rightarrow \mathcal{H}_k$ (where $\mathcal{H}_k$ is a Hilbert space), and a set of coloring functions $\mathcal{F}_{c}=\{f^{0}, f^{1}, \ldots, f^{i}\}$ on $\chi$ (with $f^{i}: \chi \rightarrow \chi$), we define the set of \textbf{iterative graph kernels (IGK)} as:
   $$
\mathbb{K}_{\mathcal{F}_{c},\phi}(x, y,i)=\langle \phi( f^{i} \circ \cdots f^{1}(x)), \phi(f^{i} \circ \cdots f^{1}(y) )\rangle=\langle\psi^{i}(x), \psi^{i}(y)\rangle
   $$
   where $x, y \in \chi$ and $\psi^{i}(\cdot)$ represents a composite function given by: $\psi^{i}=f^i \circ$ $\cdots \circ f^1$,
   Then the normalized kernel is given by:
$$
\tilde{\mathbb{K}}_{\mathcal{F}_c, \phi}(x, y, i)=\frac{\mathbb{K}_{\mathcal{F}_c, \phi}(x, y, i)}{\sqrt{\mathbb{K}_{\mathcal{F}_c, \phi}(x, x, i)} \sqrt{\mathbb{K}_{\mathcal{F}_c, \phi}(y, y, i)}}
$$
\end{definition}
Based on this definition, we can see that graph kernels utilizing the Weisfeiler-Lehman framework, including the WL-subtree kernel~\citep{DBLP:journals/jmlr/ShervashidzeSLMB11}, WLOA~\citep{DBLP:conf/nips/KriegeGW16}, and the Wasserstein Weisfeiler-Lehman (WWL) kernel~\citep{DBLP:conf/nips/TogninalliGLRB19}, should be classified as iterative graph kernels. Conversely,  the subgraph-pattern approaches, such as the graphlet kernel~\citep{DBLP:journals/jmlr/ShervashidzeVPMB09} and the shortest-path kernel~\citep{DBLP:conf/icdm/BorgwardtK05}, do not fall into this category.
\subsection{Key Properties: Monotonic Decrease \& Order Consistency}
\vspace{-0.1cm}
To effectively capture relational structures, we design the IGKs to progressively differentiate between graphs. With each iteration, additional structural features are considered, enabling the distinction of graphs that may have been indistinguishable in earlier iterations. This implies two properties: (1) the kernel values monotonically decrease with larger iterations, as the similarity between two graphs decreases with the consideration of more features; and (2) the similarity rankings across different iterations should remain consistent, meaning that graphs deemed dissimilar in early iterations should not be considered similar in later iterations.  We then provide formal definitions for these two properties and prove how they can improve generalization in the binary classification task.

\vspace{-5pt}
\begin{definition}[Monotonic Decrease]
The normalized iterative graph kernels $\tilde{\mathbb{K}}_{\mathcal{F}_{c},\phi}(x, y, i)$ are said to be monotonically decreasing if and only if:
    $$
    \tilde{\mathbb{K}}_{\mathcal{F}_{c},\phi}(x, y, i)\geq \tilde{\mathbb{K}}_{\mathcal{F}_{c},\phi}(x, y, i+1) \quad \forall x,y \in \chi
    $$
    \label{def:mono}
\end{definition}
\vspace{-0.35cm}
\begin{definition}[Order Consistency]
The normalized iterative graph kernels  $\tilde{\mathbb{K}}_{\mathcal{F}_{c},\phi}(x, y, i)$ are said to preserve order consistency if the similarity ranking remains consistent across different iterations for any pair of graphs, which is defined as:
   $$
   \tilde{\mathbb{K}}_{\mathcal{F}_{c},\phi}(x, y, i)>\tilde{\mathbb{K}}_{\mathcal{F}_{c},\phi}(x, z, i)\Rightarrow \tilde{\mathbb{K}}_{\mathcal{F}_{c},\phi}(x, y, i+1) \geq \tilde{\mathbb{K}}_{\mathcal{F}_{c},\phi}(x, z, i+1)\quad \forall x,y,z \in \chi
   $$
   \label{def:order-preserve}
\end{definition}
\vspace{-0.35cm}
Next, we show that these two properties can lead to a monotonically decreasing upper bound on the test error in the binary classification task, which suggests improved performance.

Consider a binary graph classification task and assume that the graph representations obtained at any iteration have a uniform norm. This can be achieved by simply normalizing the graph representations at the end of each iteration. That is, for any graph $x$: $\lVert\phi(\psi^i(x))\rVert = 1$. Then, for any IGK that is monotonically decreasing and preserves the order consistency, the following theorem holds: 

\begin{theorem} Let $\tilde{\mathbb{K}}_{\mathcal{F}_c, \phi}(x, y, i)$ be a normalized iterative graph kernel. Its generalization ability on the test set is provably strong when it decreases monotonically and preserves order consistency.
\label{theorem: main}
\end{theorem}

\textbf{Proof sketch.} We consider a binary classification task where supervision is provided through labels and similarity orders. We demonstrate that the presence of these two properties enhances the generalizability of the learned iterative kernel. The proof is structured in two steps: 
\begin{enumerate}
    \item [(1)] We first show that iterative kernels can learn the correct ranking given sufficient data during training, using the PAC-learning framework.
    \item[(2)] Next, we demonstrate that, once the correct ranking is learned, an iterative kernel learned with the properties of monotonic decrease and order consistency guarantees that the error rate of the graphs in the test set decreases as the number of iterations increases.
\end{enumerate}
Details of the proof can be found in Appendix~\ref{proof:generalization}.

\subsection{Theoretical Verification with WL-based Kernels }
As discussed in Section \ref{section:definition_kernel},  WL-based kernels can be categorized as iterative graph kernels, as they are generated by the coloring functions in an iterative refinement process. Consequently, a natural question arises regarding how various kernels adhere to these properties and whether their adherence reflects their actual performance. We thus investigate two popular WL-based Kernels: the WL-subtree kernel~\citep{DBLP:journals/jmlr/ShervashidzeSLMB11} and the WLOA kernel~\citep{DBLP:conf/nips/KriegeGW16}. 
\begin{theorem}
The normalized WL-subtree kernel is neither monotonically decreasing nor does it preserve order consistency.
\label{thm:WL}
\end{theorem}
\begin{theorem}
The normalized WLOA kernel is monotonically decreasing and asymptotically preserves order consistency  
when $\omega(i)=1$.
\label{thm:WLOA}
\end{theorem}
\textbf{Proof sketch.} For Theorem~\ref{thm:WL}, we illustrate a counterexample, while for Theorem~\ref{thm:WLOA}, we consider two graph pairs where the similarity condition holds:

$$
\tilde{\mathbb{K}}_{W L O A}^{(h)}\left(\mathcal{G}, \mathcal{G}^{\prime}\right) \geq \tilde{\mathbb{K}}_{W L O A}^{(h)}\left(\mathcal{G}, \mathcal{G}^{\prime \prime}\right)
$$

The similarity at the next iteration $h+1$ is scaled by a factor dependent $\omega(i), i=1, \cdots, h+1$. 
The unnormalized kernel increases monotonically, though with diminishing increments over iterations. Given this, when $\omega(i) = 1$ and $h \rightarrow \infty$, we obtain:

$$
\tilde{\mathbb{K}}_{W L O A}^{(h+1)}\left(\mathcal{G}, \mathcal{G}^{\prime}\right) \geq \tilde{\mathbb{K}}_{W L O A}^{(h+1)}\left(\mathcal{G}, \mathcal{G}^{\prime \prime}\right)
$$

We include the complete proof in Appendix~\ref{proof:WL} and~\ref{proof:WLOA}.

These findings imply that the WLOA kernel better preserves relational structures compared to the WL-subtree kernel, 
leading to improved classification performance, as supported by the literature~\citep{DBLP:conf/nips/KriegeGW16}.

\vspace{-5pt}
\section{Proposed Strategy}
\vspace{-5pt}
Given the analogy between WL-based kernels and GNNs \citep{DBLP:journals/jmlr/ShervashidzeSLMB11,DBLP:conf/icml/GilmerSRVD17}, and the observation that GNNs often fail to preserve relational structures, we hypothesize that the consistency principle is also beneficial to GNN learning.
Thus, we aim to explore how to effectively preserve this principle within the GNN architectures. 
\vspace{-0.4cm}
\begin{figure}[h]
    \centering
    \includegraphics[width=0.92\textwidth]{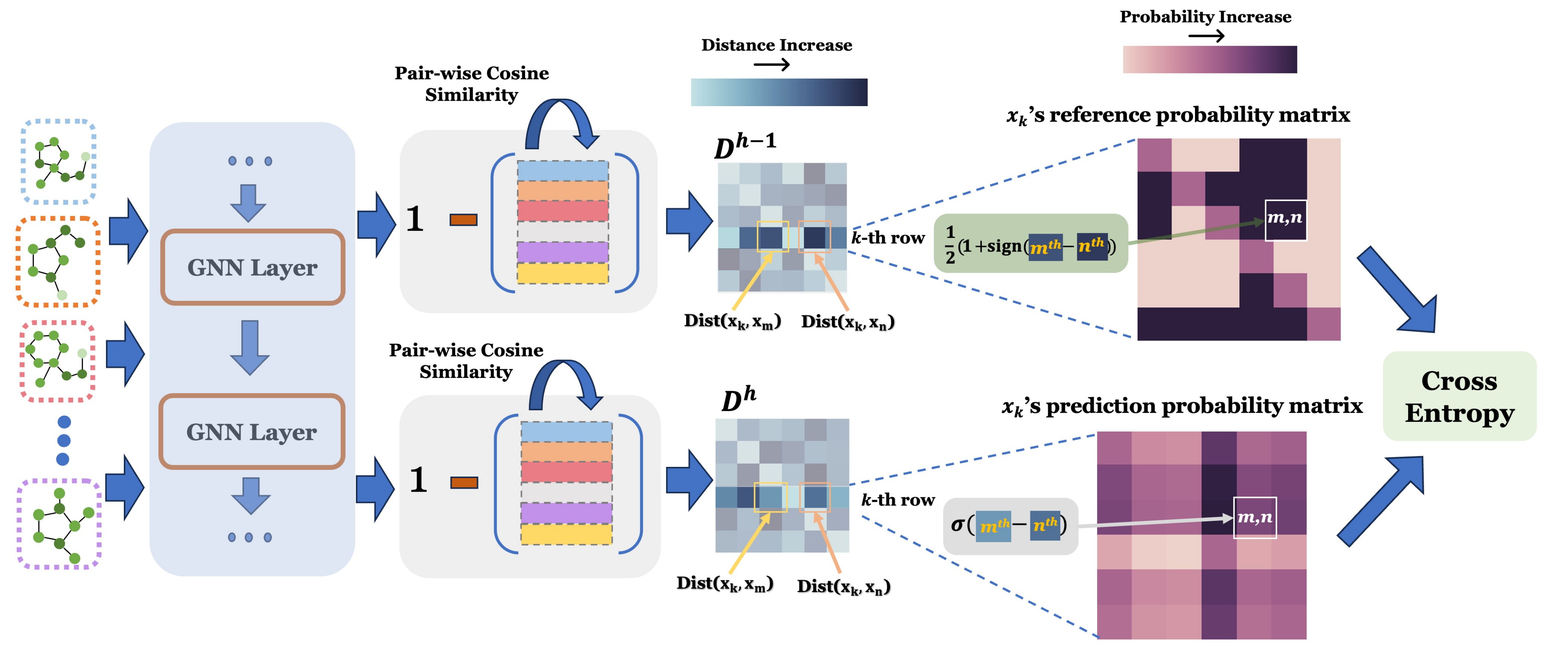}
    \vspace{-0.3cm}
    \caption{ \small Computation of Consistency loss. At each layer, pairwise distance matrix $\mathbf{D}$ is calculated using the normalized representations of graphs in a batch. After randomly selecting a reference graph $x_k$, the reference probability matrix is computed using the distance matrix from previous layer, where entry $(n,m)$ represents the known probability that the graph $x_k$ is more similar to the graph $x_n$ than to the graph $x_m$.
    For the distance matrix of current layer, we compute the predicted probability that $x_k$ is closer to $x_n$ than to $x_m$ and form the prediction probability matrix. Consistency loss is computed as the cross-entropy between the predicted and reference probability matrices}
    \vspace{-0.25cm}
    \label{fig:loss_overview}
\end{figure}

\subsection{Consistency Loss} \label{subsection:loss}
\vspace{-3pt}
Our objective is to enhance graph representation consistency across GNN layers, which has significant potential to preserve the relational structure in the representation space. If we compare $\phi(\psi^i(G))$ to the graph representations obtained at the $i$-th layer, preserving the consistency principle is equivalent to preserving the ordering of cosine similarity among the graph representations. However, due to the non-differentiable nature of ranking operations, directly minimizing the ranking differences between consecutive layers is not feasible using gradient-based optimization techniques.
 Therefore, we aim to optimize pairwise ordering relations instead of the entire ranking list.  In this work, our proposed loss employs a probabilistic approach inspired by \citep{DBLP:conf/icml/BurgesSRLDHH05}. The entire framework is illustrated in Figure \ref{fig:loss_overview}.
 
Let $\mathbf{H}^{h} \in \mathbb{R}^{n \times d}$ denote the graph embedding matrix for $n$ examples in a batch, each with a $d$-dimensional feature vector.
We first compute the distance matrix $\mathbf{D}^h$ 
for all the graphs in a batch at the $h$-th layer.
The entries ${D}_{i, j}^h$ of this matrix represent the distance between the representations of the $i$-th and $j$-th graphs, calculated as ${D}_{i, j}^h=\texttt{Dist}\left({H}_{x_i}^h, {H}_{x_j}^h\right)$. 
Here, we use the cosine distance, the complement of cosine similarity in positive space, expressed as: $1-\frac{{H}_{x_i}^h \cdot {H}_{x_j}^h}{\left\|{H}_{x_i}^h\right\| \cdot\left\|{H}_{x_j}^h\right\|}$.
Considering the distance relationship to an arbitrary graph $x_k$ in the batch, the predicted probability $\hat{\mathbb{P}}_{n, m|k}^h$ that $x_k$ is more similar to graph $x_n$ than to graph $x_m$ at layer $h$ is defined as $\hat{\mathbb{P}}_{n, m|k}^h\left(\hat{{D}}_{k, n}^h<\right.$ $\left.\hat{{D}}_{k, m}^h\right)$. This probability score, which ranges from 0 to 1, is formulated using the sigmoid function as follows:
$$
\hat{\mathbb{P}}_{n, m|k}^{h}\left(\hat{{D}}_{k, n}^{h}< \hat{{D}}_{k, m}^{h}\right)=\frac{1}{1+\exp \left(\hat{{D}}_{k, m}^h-\hat{{D}}_{k, n}^h\right)}
$$

Given the distance matrix from the previous layer ${D}^{h-1}$, the known probability that graph $x_k$ is more similar to graph $x_n$ than to graph $x_m$, denoted as  $\tilde{\mathbb{P}}_{n, m}^{h-1}(\hat{{D}}_{k, n}^{h-1}, \hat{{D}}_{k, m}^{h-1})$,  can be formulated as follows:
\begin{align*}
\tilde{\mathbb{P}}_{n, m|k}^{h-1} = \frac{1}{2} \left(1 + \operatorname{sign}(\hat{D}_{k, n}^{h-1} - \hat{D}_{k, m}^{h-1})\right)
\end{align*}

We can then minimize the discrepancy between the predicted and the known probability distributions  to enhance representation consistency across the layers.
Here,we employ the cross-entropy loss to effectively measure the divergence between these two distributions. Specifically, for a pair $\left(x_n, x_m\right)$ centered on $x_k$ at layer $h$, the cross-entropy loss can be expressed as:
$$
\mathcal{L}_{\text {cross-entropy}}\left((x_n, x_m)\mid x_k, h\right)=-\tilde{\mathbb{P}}_{n, m|k}^{h-1} \log \hat{\mathbb{P}}_{n, m|k}^{h}-\left(1- \tilde{\mathbb{P}}_{n, m|k}^{h-1}\right) \log \left(1-\hat{\mathbb{P}}_{n, m|k}^{h}\right)
$$

Then, the total loss function, which quantifies the consistency of pair-wise distance relations for graph $x_k$ at layer $h$, can be formulated as
$$
\mathcal{L}_{\text {consistency }}(k)=\sum_{n, m} \mathcal{L}((x_n, x_m)\mid x_k, h)
$$
The overall objective function of the proposed framework can then be formulated as the weighted sum of the original loss and the consistency loss.
$$
\mathcal{L}_{\text {total }}=\mathcal{L}_{\text {origin}}+\lambda \sum_{i} \mathcal{L}_{\text {consistency }}(i)
$$

Here, $\lambda$ is a hyperparameter that controls the strength of the consistency constraint.
\vspace{-5pt}

\section{Experiment}
\label{sec:experiment}
In this section, we examine the effectiveness 
of the proposed consistency loss for the graph classification task. Specifically, we aim to address the following questions:
\textbf{Q1:} Does the consistency loss effectively enhance the performance of 
various GNN backbones in the graph classification task? 
\textbf{Q2:} How does the 
consistency loss influence the rank correlation 
of graph similarities across GNN layers?
\textbf{Q3:} How does the consistency loss influence dataset performance across varying levels of complexity, both in structural intricacy and task difficulty?
\subsection{Experiment Setup}
\vspace{-2pt}
\paragraph{Dataset} We conduct extensive experiments using the TU Dataset \citep{DBLP:journals/corr/abs-2007-08663}, the Open Graph Benchmark (OGB) \citep{DBLP:conf/nips/HuFZDRLCL20} and Reddit Threads(Reddit-T) dataset \citep{DBLP:conf/log/BauseK22}. The TU Dataset consists of eight graph classification datasets, categorized into three main types: (1) Chem/Bioinformatics datasets, including D\&D, NCI1, NCI109, and PROTEINS; (2) Social Network datasets, including IMDB-BINARY,IMDB-MULTI and COLLAB, where a constant feature value of 1 was assigned to all vertices due to the absence of vertex features; and (3) a Computer Vision dataset, COIL-RAG. The OGB datasets include ogbg-molhiv, a molecular property prediction dataset used to determine whether a molecule inhibits HIV replication.  Reddit-T dataset is used for classifying threads from Reddit as either discussions or non-discussions, where users are represented as nodes and replies between them as links.

For the TU dataset and Reddit-T, consistent with prior work \citep{DBLP:conf/iclr/XinyiC19,DBLP:conf/iclr/XuHLJ19}, we utilize an 8:1:1 ratio for training, validation, and testing sets. For the OGB datasets, we use the official splits provided. 
Training stops at the highest validation performance, and test accuracy is taken from the corresponding epoch in each fold. Final results are reported as the mean accuracy (except {ogbg-molhiv}) and standard deviation over $10$ folds. 
For {ogbg-molhiv}, we follow the official evaluator and use ROC-AUC as the evaluation metric. 
Detailed information about these datasets can be found in Appendix~\ref{app:dataset}.
\vspace{-4pt}
\paragraph{Model} We use three widely adopted GNN models as baselines: namely, GCN \citep{DBLP:conf/iclr/KipfW17}, GIN \citep{DBLP:conf/iclr/XuHLJ19}, and GraphSAGE \citep{DBLP:conf/nips/HamiltonYL17}. We also include two recent GNN models, namely  GTransformer \citep{DBLP:conf/ijcai/ShiHFZWS21} and GMT \citep{DBLP:conf/iclr/BaekKH21}. To ensure a fair comparison, we maintain the same number of layers and layer sizes for both the base models and the models with our proposed consistency loss, ensuring the sharing of the same network architecture. Detailed information about hyperparameter tuning is provided in Appendix~\ref{app:exp_setup}.
\subsection{Effectiveness of Consistency Loss}
\vspace{-3pt}
To answer \textbf{Q1}, we present the results for the TU, OGB and Reddit-T datasets in Table \ref{table:tu}. As shown in this table, GNN models with the consistency loss yield significant performance improvements over their base models on different datasets. These findings suggest that the consistency framework is a versatile and robust approach for enhancing the predictive capabilities of GNNs in real-world datasets, irrespective of the base model and dataset domain. Notably, the GIN method demonstrates the most significant improvements, achieving enhancements of up to 4.51\% on the D\&D dataset, 4.32\% on the COLLAB dataset, and 3.70\% on the IMDB-B dataset. This improvement can be linked to our empirical observation regarding the weak ability of GIN to preserve consistency across layers. In addition, our method demonstrates satisfactory improvements on datasets with numerous classes (e.g., COIL-RAG) and large-scale datasets (e.g., ogbg-molhiv and Reddit-T), indicating that our approach is both flexible and scalable for handling complex and extensive datasets.
\vspace{-7pt}
\begin{table}[h]
\setlength{\tabcolsep}{2.0pt}
\caption{\small 
Classification performance on the TU, OGB and Reddit-T datasets, with and without the consistency loss.  Highlighted cells indicate instances where the base GNN with the consistency loss outperforms the base GNN alone. 
 The reported values are average accuracy for TU datasets and ROC-AUC for the ogbg-molhiv dataset, including their standard deviations.}
 \begin{adjustbox}{width=\textwidth,keepaspectratio}
\begin{tabular}{lcccccccccc}
\toprule\toprule
& NCI1            & NCI109          & PROTEINS        & D\&D            & IMDB-B            & IMDB-M      & COLLAB  & COIL-RAG  & OGB-HIV & REDDIT-T \\
\#Graphs             & 4110            & 4127            & 1113            & 1178            & 1000              & 1500           & 5000   & 3900 & 41127  & 203088 \\ 
Avg. \#nodes         & 29.87           & 29.68           & 39.06           & 284.32          & 19.77             & 13.00          & 74.49  & 3.01 & 25.50   & 23.93 \\ 
\toprule
GCN                  & 73.96\tiny{$\pm$} 2.37 & 74.04\tiny{$\pm$} 3.09 & 73.24\tiny{$\pm$} 6.93 & 74.92\tiny{$\pm$} 2.66 & 75.40\tiny{$\pm$} 2.97   & 55.07\tiny{$\pm$}1.24 & 81.72\tiny{$\pm$}0.84 & 91.72\tiny{$\pm$}1.65 & 72.86\tiny{$\pm$}1.90 & 76.00\tiny{$\pm$}0.44 \\
\quad +$\mathcal{L}_{\text{consistency}}$     &\empcell 75.12\tiny{$\pm$} 1.19 &  73.25\tiny{$\pm$} 1.25 &\empcell  75.07\tiny{$\pm$} 5.05 & \empcell78.56\tiny{$\pm$} 3.32 &\empcell 75.85\tiny{$\pm$} 1.82   &\empcell 56.27\tiny{$\pm$}1.00 &\empcell 83.44\tiny{$\pm$}0.45 &\empcell 93.38\tiny{$\pm$}1.64 &\empcell 73.75\tiny{$\pm$}0.89 &\empcell 77.12\tiny{$\pm$}0.12 \\ 
\midrule
GIN                  & 78.13\tiny{$\pm$} 2.11 & 76.75\tiny{$\pm$} 2.91 & 72.97\tiny{$\pm$} 4.59 & 71.10\tiny{$\pm$} 4.63 & 70.80\tiny{$\pm$} 4.07   & 52.13\tiny{$\pm$}1.42 & 79.84\tiny{$\pm$}1.05 & 93.33\tiny{$\pm$}1.48 &  71.60 \tiny{$\pm$}2.36 & 77.50\tiny{$\pm$}0.16 \\
\quad +$\mathcal{L}_{\text{consistency}}$    &\empcell 79.45\tiny{$\pm$} 1.09 &\empcell 77.46\tiny{$\pm$} 1.96 & \empcell 74.98\tiny{$\pm$} 4.57 &\empcell 75.51\tiny{$\pm$} 2.63 &\empcell 74.50\tiny{$\pm$} 3.06   &\empcell 53.46\tiny{$\pm$}2.44 &\empcell 84.16\tiny{$\pm$}0.81 &\empcell 94.03\tiny{$\pm$}1.33 &\empcell 74.57\tiny{$\pm$}1.61 & \empcell 77.64\tiny{$\pm$}0.05 \\ 
\midrule
GraphSAGE            & 74.40\tiny{$\pm$} 1.83 & 73.17\tiny{$\pm$} 0.47 & 74.96\tiny{$\pm$} 3.14  & 76.44\tiny{$\pm$} 4.16 & 73.90\tiny{$\pm$} 2.17 & 51.33\tiny{$\pm$}2.95 & 78.92\tiny{$\pm$}1.20 & 89.56\tiny{$\pm$}2.37 & 77.03\tiny{$\pm$}1.65 & 76.67\tiny{$\pm$}0.11 \\
\quad +$\mathcal{L}_{\text{consistency}}$     & \empcell 78.26\tiny{$\pm$} 1.08 &\empcell 74.10\tiny{$\pm$} 2.10 &\empcell 76.40\tiny{$\pm$} 3.12 &\empcell 77.50\tiny{$\pm$} 3.38 &\empcell 74.75\tiny{$\pm$} 3.06   & \empcell 54.27\tiny{$\pm$}1.24 &\empcell  82.12\tiny{$\pm$}0.78 &\empcell 92.31\tiny{$\pm$}1.32 &\empcell  78.60\tiny{$\pm$}1.44 &\empcell 77.57\tiny{$\pm$}0.05 \\ 
\midrule
GTransformer         & 75.72\tiny{$\pm$}2.69  & 74.79\tiny{$\pm$}1.82  & 73.33\tiny{$\pm$}4.80    & 75.42\tiny{$\pm$}3.22   & 72.20\tiny{$\pm$}3.49   & 53.33\tiny{$\pm$}1.12 & 80.36\tiny{$\pm$}0.56 & 83.74\tiny{$\pm$}3.17 & 76.81\tiny{$\pm$}1.34 & 76.75\tiny{$\pm$}0.12 \\
\quad +$\mathcal{L}_{\text{consistency}}$   & \empcell 76.83\tiny{$\pm$}1.36  & \empcell 75.82\tiny{$\pm$}1.53  & \empcell 77.03\tiny{$\pm$}3.79   & \empcell 76.57\tiny{$\pm$}2.54 &\empcell 73.75\tiny{$\pm$}2.56 &\empcell 56.53\tiny{$\pm$}1.54 &\empcell 80.48\tiny{$\pm$}0.47&\empcell 91.67\tiny{$\pm$}1.88 &\empcell 76.90\tiny{$\pm$}3.25 &  \empcell 77.14\tiny{$\pm$}0.06 \\ 
\midrule
GMT         & 75.04\tiny{$\pm$}1.43   & 73.90\tiny{$\pm$}2.29   & 72.70\tiny{$\pm$}4.21  & 72.80\tiny{$\pm$}2.19 & 79.80\tiny{$\pm$}1.08    & 54.13\tiny{$\pm$}2.90 & 80.36\tiny{$\pm$}1.15 & 90.85\tiny{$\pm$}1.91 & 74.86\tiny{$\pm$}2.26 & 72.06\tiny{$\pm$}10.15 \\
\quad +$\mathcal{L}_{\text{consistency}}$    &\empcell 75.52\tiny{$\pm$}1.07    &\empcell 75.20\tiny{$\pm$}0.95  & \empcell 74.86\tiny{$\pm$}2.03    &\empcell 73.14\tiny{$\pm$}2.28 & 79.60\tiny{$\pm$}1.91 &\empcell 54.80\tiny{$\pm$}1.42 & \empcell 82.80\tiny{$\pm$}0.61&\empcell 92.00\tiny{$\pm$}1.43 &\empcell 76.00\tiny{$\pm$}1.99 &\empcell 77.19\tiny{$\pm$}0.14 \\  
\bottomrule
\end{tabular}
\end{adjustbox}
\label{table:tu}
\end{table}
\vspace{-5pt}

Furthermore, we analyze the complexity and scalability of our method on the TU Dataset, with additional details provided in Appendices~\ref{app:timecomplexity} and~\ref{app:spacecomplexity}. Additionally, we demonstrate the method's potential to enhance performance significantly, even with a marginal increase in computational cost, as illustrated in Appendix~\ref{app:efficiency}.
\vspace{-5pt}
\subsection{Effect of the Consistency Loss on Rank Correlation }
\begin{wraptable}{R}{0.49\textwidth}
\large
\setlength{\tabcolsep}{5pt}
\resizebox{0.49\textwidth}{!}{ 
\begin{minipage}[t]{0.50\textwidth} 
\caption{\small Spearman correlation was computed for graph representations from consecutive layers on the TU datasets, both with and without consistency loss. Values with higher rank correlation are highlighted. The consistency loss can enhance the rank correlation of graph similarities.}
\begin{adjustbox}{width=\textwidth,keepaspectratio}
\setlength{\tabcolsep}{3pt}
\begin{tabular}{lcccccc}
    \toprule\toprule
    & NCI1            & NCI109          & PROTEINS        & D\&D            & IMDB-B                   \\
    \toprule
    GCN                  & 0.753 & 0.920 & 0.584 & 0.709 & 0.846              \\
    \quad +$\mathcal{L}_{\text{consistency}}$    & \empcell 0.859 & \empcell 0.958 & \empcell 0.946 & \empcell 0.896 & \empcell 0.907               \\ 
    \midrule
    GIN                  & 0.666 & 0.674 & 0.741 & 0.721 & 0.598     \\
    \quad +$\mathcal{L}_{\text{consistency}}$     & \empcell 0.877 & \empcell 0.821 & \empcell 0.904 & \empcell 0.847 & \empcell 0.816               \\ 
    \midrule
    GraphSAGE            & 0.903 & 0.504 & 0.845 & 0.741 & 0.806  \\
    \quad +$\mathcal{L}_{\text{consistency}}$    & \empcell 0.911 & \empcell 0.709 & \empcell 0.916 & \empcell 0.872 & \empcell 0.933               \\ 
    \midrule
    GTransformer         & 0.829  & 0.817  & 0.867    & 0.865   & 0.884        \\
    \quad +$\mathcal{L}_{\text{consistency}}$     & \empcell 0.863  & \empcell 0.883  & \empcell 0.915   & \empcell 0.880   & \empcell 0.917                 \\ 
    \midrule
    GMT         &  0.872   &  0.887  &  0.980  &   0.826  &    0.893    \\
    \quad +$\mathcal{L}_{\text{consistency}}$     & \empcell 0.906 & \empcell  0.908 & \empcell 0.983  & \empcell 0.856 & \empcell 0.908                 \\ 
    \bottomrule
    \end{tabular}
\end{adjustbox}
\label{table:tu_rank}
\end{minipage}
}
\vspace{-15pt}
\end{wraptable}
To answer \textbf{Q2}, we compare the consistency of graph representations across layers with and without the proposed consistency loss. Specifically, we use the Spearman's rank correlation coefficient, a widely accepted method for computing correlation between ranked variables, to quantitatively measure the consistency of graph similarities across layers.  For a fair comparison, we construct a distance matrix $D^h$ for all test data at each layer $h$, where each row $D_{x_i ;}^h$ represents the distances from graph $x_i$ to all other graphs. We then compute the rank correlation between $D_{x_i,:}^h$ and $D_{x_i ;}^{h+1}$ for each graph $x_i$.

We average the correlation values for all graphs to obtain the overall correlation for layer $h$. Then, we compute the mean of these values across layers, enabling a global comparison of relational consistency throughout the model and dataset. All results were averaged over 5 repeated experiments with same training setting.

We present our results on a series of datasets from the TU Dataset in Table \ref{table:tu_rank}.  As shown in the table, it is evident that the representation space becomes more consistent with our proposed consistency loss. For example, a significant enhancement is observed for the GIN model. Another notable point is the result for the GCN model on the NCI109 dataset and for the GMT model on the IMDB-B dataset. We find that the correlation is already fairly high even without the implementation of $\mathcal{L}_{\text{consistency}}$, resulting in minimal correlation improvements with our method. This phenomenon provides a plausible explanation for why our method is not effective in these two cases.
\subsection{{Study on Task Complexity} }
To address \textbf{Q3} and further evaluate the performance of our method across different scenarios, we extended our study by conducting experiments on graph datasets with increasing task and structural complexity.
\vspace{-5pt}
\paragraph{Increasing Task Complexity}
We increase the task complexity by expanding the number of classes that the model needs to classify.
To assess the effect of increased class complexity on our method’s performance, we sampled subsets from the REDDIT-MULTI 5K dataset \citep{DBLP:conf/kdd/YanardagV15} with a progressively greater number of classes, which originally consists of five classes. Specifically, we randomly sampled between 2 and 4 classes to construct new datasets from the original dataset and conducted classification tasks using both GCN and GCN with $\mathcal{L}_{\text{consistency}}$ on these newly constructed datasets. We report the mean test accuracy over five experiments for each subset, as presented in Table \ref{table:subset_performance}.
\vspace{-12pt}
\begin{table}[h!]
\centering
\caption{Performance comparison across different subsets and the full set.}
\label{table:subset_performance}
\begin{tabular}{ccccc}
\toprule
 & \textbf{Subset1} & \textbf{Subset2} & \textbf{Subset3} & \textbf{Fullset} \\
 & \textbf{(2 classes)} & \textbf{(3 classes)} & \textbf{(4 classes)} & \textbf{(5 classes)} \\ \midrule
GCN  & 79.50 & 67.13 & 50.30 & 53.80 \\
GCN+$\mathcal{L}_{\text{consistency}}$ & 81.10 & 68.00 & 57.15 & 57.12 \\
\bottomrule
\end{tabular}
\end{table}
\vspace{-2pt}

The results demonstrate that the effectiveness of our method remains robust, even as the number of classes increases. In fact, it may provide greater advantages when applied to multi-class classification tasks. This resilience likely stems from our method’s focus on identifying relational structures in the intermediate representations of GNN models, rather than relying heavily on label information. This approach helps mitigate the impact of potential label noise in the original data.
These findings align with the noticeable performance improvements observed in both binary and multi-class classification tasks, as shown in Table~\ref{table:tu}.
\vspace{-2pt}
\paragraph{Increasing Structural Complexity}
We assessed the impact of structural complexity by partitioning the IMDB-B dataset into three subsets with progressively increasing graph densities. Graph density, denoted as $d=\frac{2M}{N(N-1)}$, where $N$ is the number of nodes and $M$ is the number of edges in graph $\mathcal{G}$, was used as the criterion for creating these subsets. The dataset was divided into three groups: (small) for graphs with densities below the 33rd percentile, (median) for densities between the 33rd and 67th percentiles, and (large) for graphs with densities above the 67th percentile. We applied both GCN and GCN+$\mathcal{L}_{\text{consistency}}$ models to these subsets, and the results are summarized in Table \ref{table:imdb_b_performance}.
\vspace{-3pt}
\begin{table}[h!]
\centering
\caption{Performance comparison on IMDB-B datasets of different densities.}
\label{table:imdb_b_performance}
\begin{tabular}{cccc}
\toprule
 & IMDB-B & IMDB-B & IMDB-B \\
 & \textbf{(small)} & \textbf{(medium)} & \textbf{(large)} \\ \midrule
GCN & 77.58\tiny{$\pm 4.11$} & 66.25\tiny{$\pm 5.38$} & 67.61\tiny{$\pm 6.21$} \\
GCN+$\mathcal{L}_{\text{consistency}}$ & 84.24\tiny{$\pm 4.85$} & 69.06\tiny{$\pm 4.06$} & 71.43\tiny{$\pm 4.43$} \\
\bottomrule
\end{tabular}
\end{table}
\par The above results show that the GCN model, enhanced with the $\mathcal{L}_{\text{consistency}}$ loss function, consistently outperforms the original version across different structural complexity groups, demonstrating the robustness and effectiveness of the proposed method. 

Furthermore, we evaluate the method's efficiency under various complexity scenarios, as detailed in the Appendix~\ref{app:taskcomplexity}.
\vspace{-10pt}
\vspace{-2pt}
\section{Related Work}
\vspace{-10pt}
\paragraph{Graph Distance and Similarity}
Measuring distances or similarities between graphs is a fundamental problem in graph learning. Graph kernels, which define graph similarity, have gained significant attention. Most graph kernels use the $\mathcal{R}$-Convolution framework \citep{haussler1999convolution} to compare substructure similarities. A trailblazing kernel by \citep{DBLP:conf/icml/KashimaTI03} used node and edge attributes to generate label sequences through a random walk.  
The WL-subtree kernel \citep{DBLP:journals/jmlr/ShervashidzeSLMB11} generates graph-level features by summing node representation contributions. Recent works align matching substructures between graphs. For instance, \citet{DBLP:conf/nips/KriegeGW16} proposed a discrete optimal assignment kernel based on vertex kernels from WL labels. \citet{DBLP:conf/nips/TogninalliGLRB19} extended this to include fractional assignments using the Wasserstein distance. Additionally, measuring graph distances is also a prevalent problem. \citet{DBLP:conf/icml/VayerCTCF19} combined the Wasserstein and Gromov-Wasserstein distances \citep{villani2003topics, DBLP:journals/focm/Memoli11}. \citet{DBLP:conf/icml/ChenLMWW22} proposed a polynomial-time WL distance for labeled Markov chains, treating labeled graphs as a special case. 
\vspace{-5pt}
\paragraph{Bridging Graph Kernels and GNNs}
Many studies have explored the connection between graph kernels and GNNs, attempting to integrate them into a unified framework. Certain approaches focus on leveraging GNN architecture to design novel kernels. For instance, \citet{DBLP:conf/nips/MairalKHS14} presents neural network architectures that learn graph representations within the Reproducing Kernel Hilbert Space (RKHS) of graph kernels. Similarly, \citet{DBLP:conf/nips/DuHSPWX19} proposed a graph kernel equivalent to infinitely wide GNNs, which can be trained using gradient descent. Conversely, other studies have incorporated kernel methods directly into GNNs. For example, \citet{DBLP:conf/nips/NikolentzosV20} utilize graph kernels as convolutional filters within GNN architectures. Additionally, \citet{DBLP:conf/www/LeeZA24} proposes a novel Kernel Convolution Network that employs the random walk kernel as the core mechanism for learning descriptive graph features. Instead of applying specific kernel patterns as mentioned in previous work, we introduce a general method for GNNs to capture consistent similarity relationships, thereby enhancing classification performance.
\vspace{-8pt}
\section{Conclusion}
\vspace{-10pt}
In this paper, we study a class of graph kernels and introduce the concept of consistency property in graph classification tasks. We theoretically prove that this property leads to a more structure-aware representation space for classification using kernel methods. Based on this analysis, we extend this principle to enhance GNN models. We propose a novel, model-agnostic consistency learning framework for GNNs that enables them to capture relational structures in the graph representation space.
Experiments show that our proposed method universally enhances the performance of backbone networks on graph classification benchmarks, providing new insights into bridging the gap between traditional kernel methods and GNN models.

\bibliographystyle{plainnat}
\bibliography{ref}

\begin{thebibliography}{45}
\providecommand{\natexlab}[1]{#1}
\providecommand{\url}[1]{\texttt{#1}}
\expandafter\ifx\csname urlstyle\endcsname\relax
  \providecommand{\doi}[1]{doi: #1}\else
  \providecommand{\doi}{doi: \begingroup \urlstyle{rm}\Url}\fi

\bibitem[Baek et~al.(2021)Baek, Kang, and Hwang]{DBLP:conf/iclr/BaekKH21}
Jinheon Baek, Minki Kang, and Sung~Ju Hwang.
\newblock Accurate learning of graph representations with graph multiset pooling.
\newblock In \emph{9th International Conference on Learning Representations, {ICLR} 2021, Virtual Event, Austria, May 3-7, 2021}. OpenReview.net, 2021.
\newblock URL \url{https://openreview.net/forum?id=JHcqXGaqiGn}.

\bibitem[Bause and Kriege(2022)]{DBLP:conf/log/BauseK22}
Franka Bause and Nils~Morten Kriege.
\newblock Gradual weisfeiler-leman: Slow and steady wins the race.
\newblock In Bastian Rieck and Razvan Pascanu, editors, \emph{Learning on Graphs Conference, LoG 2022, 9-12 December 2022, Virtual Event}, volume 198 of \emph{Proceedings of Machine Learning Research}, page~20. {PMLR}, 2022.
\newblock URL \url{https://proceedings.mlr.press/v198/bause22a.html}.

\bibitem[Borgwardt and Kriegel(2005)]{DBLP:conf/icdm/BorgwardtK05}
Karsten~M. Borgwardt and Hans{-}Peter Kriegel.
\newblock Shortest-path kernels on graphs.
\newblock In \emph{Proceedings of the 5th {IEEE} International Conference on Data Mining {(ICDM} 2005), 27-30 November 2005, Houston, Texas, {USA}}, pages 74--81. {IEEE} Computer Society, 2005.
\newblock \doi{10.1109/ICDM.2005.132}.
\newblock URL \url{https://doi.org/10.1109/ICDM.2005.132}.

\bibitem[Burges et~al.(2005)Burges, Shaked, Renshaw, Lazier, Deeds, Hamilton, and Hullender]{DBLP:conf/icml/BurgesSRLDHH05}
Christopher J.~C. Burges, Tal Shaked, Erin Renshaw, Ari Lazier, Matt Deeds, Nicole Hamilton, and Gregory~N. Hullender.
\newblock Learning to rank using gradient descent.
\newblock In Luc~De Raedt and Stefan Wrobel, editors, \emph{Machine Learning, Proceedings of the Twenty-Second International Conference {(ICML} 2005), Bonn, Germany, August 7-11, 2005}, volume 119 of \emph{{ACM} International Conference Proceeding Series}, pages 89--96. {ACM}, 2005.
\newblock \doi{10.1145/1102351.1102363}.
\newblock URL \url{https://doi.org/10.1145/1102351.1102363}.

\bibitem[Chen et~al.(2022)Chen, Lim, M{\'{e}}moli, Wan, and Wang]{DBLP:conf/icml/ChenLMWW22}
Samantha Chen, Sunhyuk Lim, Facundo M{\'{e}}moli, Zhengchao Wan, and Yusu Wang.
\newblock Weisfeiler-lehman meets gromov-wasserstein.
\newblock In Kamalika Chaudhuri, Stefanie Jegelka, Le~Song, Csaba Szepesv{\'{a}}ri, Gang Niu, and Sivan Sabato, editors, \emph{International Conference on Machine Learning, {ICML} 2022, 17-23 July 2022, Baltimore, Maryland, {USA}}, volume 162 of \emph{Proceedings of Machine Learning Research}, pages 3371--3416. {PMLR}, 2022.
\newblock URL \url{https://proceedings.mlr.press/v162/chen22o.html}.

\bibitem[Du et~al.(2019)Du, Hou, Salakhutdinov, P{\'{o}}czos, Wang, and Xu]{DBLP:conf/nips/DuHSPWX19}
Simon~S. Du, Kangcheng Hou, Ruslan Salakhutdinov, Barnab{\'{a}}s P{\'{o}}czos, Ruosong Wang, and Keyulu Xu.
\newblock Graph neural tangent kernel: Fusing graph neural networks with graph kernels.
\newblock In Hanna~M. Wallach, Hugo Larochelle, Alina Beygelzimer, Florence d'Alch{\'{e}}{-}Buc, Emily~B. Fox, and Roman Garnett, editors, \emph{Advances in Neural Information Processing Systems 32: Annual Conference on Neural Information Processing Systems 2019, NeurIPS 2019, December 8-14, 2019, Vancouver, BC, Canada}, pages 5724--5734, 2019.
\newblock URL \url{https://proceedings.neurips.cc/paper/2019/hash/663fd3c5144fd10bd5ca6611a9a5b92d-Abstract.html}.

\bibitem[Gilmer et~al.(2017)Gilmer, Schoenholz, Riley, Vinyals, and Dahl]{DBLP:conf/icml/GilmerSRVD17}
Justin Gilmer, Samuel~S. Schoenholz, Patrick~F. Riley, Oriol Vinyals, and George~E. Dahl.
\newblock Neural message passing for quantum chemistry.
\newblock In Doina Precup and Yee~Whye Teh, editors, \emph{Proceedings of the 34th International Conference on Machine Learning, {ICML} 2017, Sydney, NSW, Australia, 6-11 August 2017}, volume~70 of \emph{Proceedings of Machine Learning Research}, pages 1263--1272. {PMLR}, 2017.
\newblock URL \url{http://proceedings.mlr.press/v70/gilmer17a.html}.

\bibitem[Hamilton et~al.(2017)Hamilton, Ying, and Leskovec]{DBLP:conf/nips/HamiltonYL17}
William~L. Hamilton, Zhitao Ying, and Jure Leskovec.
\newblock Inductive representation learning on large graphs.
\newblock In Isabelle Guyon, Ulrike von Luxburg, Samy Bengio, Hanna~M. Wallach, Rob Fergus, S.~V.~N. Vishwanathan, and Roman Garnett, editors, \emph{Advances in Neural Information Processing Systems 30: Annual Conference on Neural Information Processing Systems 2017, December 4-9, 2017, Long Beach, CA, {USA}}, pages 1024--1034, 2017.
\newblock URL \url{https://proceedings.neurips.cc/paper/2017/hash/5dd9db5e033da9c6fb5ba83c7a7ebea9-Abstract.html}.

\bibitem[Haussler et~al.(1999)]{haussler1999convolution}
David Haussler et~al.
\newblock Convolution kernels on discrete structures.
\newblock Technical report, Citeseer, 1999.

\bibitem[Hu et~al.(2020)Hu, Fey, Zitnik, Dong, Ren, Liu, Catasta, and Leskovec]{DBLP:conf/nips/HuFZDRLCL20}
Weihua Hu, Matthias Fey, Marinka Zitnik, Yuxiao Dong, Hongyu Ren, Bowen Liu, Michele Catasta, and Jure Leskovec.
\newblock Open graph benchmark: Datasets for machine learning on graphs.
\newblock In Hugo Larochelle, Marc'Aurelio Ranzato, Raia Hadsell, Maria{-}Florina Balcan, and Hsuan{-}Tien Lin, editors, \emph{Advances in Neural Information Processing Systems 33: Annual Conference on Neural Information Processing Systems 2020, NeurIPS 2020, December 6-12, 2020, virtual}, 2020.
\newblock URL \url{https://proceedings.neurips.cc/paper/2020/hash/fb60d411a5c5b72b2e7d3527cfc84fd0-Abstract.html}.

\bibitem[Huang et~al.(2024)Huang, Yang, Zhou, and Yan]{huangenhancing}
Zheng Huang, Qihui Yang, Dawei Zhou, and Yujun Yan.
\newblock Enhancing size generalization in graph neural networks through disentangled representation learning.
\newblock In \emph{Forty-first International Conference on Machine Learning}, 2024.

\bibitem[Kashima et~al.(2003)Kashima, Tsuda, and Inokuchi]{DBLP:conf/icml/KashimaTI03}
Hisashi Kashima, Koji Tsuda, and Akihiro Inokuchi.
\newblock Marginalized kernels between labeled graphs.
\newblock In Tom Fawcett and Nina Mishra, editors, \emph{Machine Learning, Proceedings of the Twentieth International Conference {(ICML} 2003), August 21-24, 2003, Washington, DC, {USA}}, pages 321--328. {AAAI} Press, 2003.
\newblock URL \url{http://www.aaai.org/Library/ICML/2003/icml03-044.php}.

\bibitem[Kipf and Welling(2017)]{DBLP:conf/iclr/KipfW17}
Thomas~N. Kipf and Max Welling.
\newblock Semi-supervised classification with graph convolutional networks.
\newblock In \emph{5th International Conference on Learning Representations, {ICLR} 2017, Toulon, France, April 24-26, 2017, Conference Track Proceedings}. OpenReview.net, 2017.
\newblock URL \url{https://openreview.net/forum?id=SJU4ayYgl}.

\bibitem[Kriege et~al.(2016)Kriege, Giscard, and Wilson]{DBLP:conf/nips/KriegeGW16}
Nils~M. Kriege, Pierre{-}Louis Giscard, and Richard~C. Wilson.
\newblock On valid optimal assignment kernels and applications to graph classification.
\newblock In Daniel~D. Lee, Masashi Sugiyama, Ulrike von Luxburg, Isabelle Guyon, and Roman Garnett, editors, \emph{Advances in Neural Information Processing Systems 29: Annual Conference on Neural Information Processing Systems 2016, December 5-10, 2016, Barcelona, Spain}, pages 1615--1623, 2016.
\newblock URL \url{https://proceedings.neurips.cc/paper/2016/hash/0efe32849d230d7f53049ddc4a4b0c60-Abstract.html}.

\bibitem[Kriege et~al.(2020)Kriege, Johansson, and Morris]{DBLP:journals/ans/KriegeJM20}
Nils~M. Kriege, Fredrik~D. Johansson, and Christopher Morris.
\newblock A survey on graph kernels.
\newblock \emph{Appl. Netw. Sci.}, 5\penalty0 (1):\penalty0 6, 2020.
\newblock \doi{10.1007/S41109-019-0195-3}.
\newblock URL \url{https://doi.org/10.1007/s41109-019-0195-3}.

\bibitem[Lee et~al.(2024)Lee, Zhao, and Akoglu]{DBLP:conf/www/LeeZA24}
Meng{-}Chieh Lee, Lingxiao Zhao, and Leman Akoglu.
\newblock Descriptive kernel convolution network with improved random walk kernel.
\newblock In Tat{-}Seng Chua, Chong{-}Wah Ngo, Ravi Kumar, Hady~W. Lauw, and Roy~Ka{-}Wei Lee, editors, \emph{Proceedings of the {ACM} on Web Conference 2024, {WWW} 2024, Singapore, May 13-17, 2024}, pages 457--468. {ACM}, 2024.
\newblock \doi{10.1145/3589334.3645405}.
\newblock URL \url{https://doi.org/10.1145/3589334.3645405}.

\bibitem[Li et~al.(2023{\natexlab{a}})Li, Duda, Zhang, Koutra, and Yan]{li2023interpretable}
Gaotang Li, Marlena Duda, Xiang Zhang, Danai Koutra, and Yujun Yan.
\newblock Interpretable sparsification of brain graphs: Better practices and effective designs for graph neural networks.
\newblock In \emph{Proceedings of the 29th ACM SIGKDD Conference on Knowledge Discovery and Data Mining}, pages 1223--1234, 2023{\natexlab{a}}.

\bibitem[Li et~al.(2023{\natexlab{b}})Li, Koutra, and Yan]{li2023size}
Gaotang Li, Danai Koutra, and Yujun Yan.
\newblock Size generalization of graph neural networks on biological data: Insights and practices from the spectral perspective.
\newblock \emph{arXiv preprint arXiv:2305.15611}, 2023{\natexlab{b}}.

\bibitem[Lin et~al.(2023)Lin, Chen, and Wang]{DBLP:conf/iclr/LinCW23}
Lu~Lin, Jinghui Chen, and Hongning Wang.
\newblock Spectral augmentation for self-supervised learning on graphs.
\newblock In \emph{The Eleventh International Conference on Learning Representations, {ICLR} 2023, Kigali, Rwanda, May 1-5, 2023}. OpenReview.net, 2023.
\newblock URL \url{https://openreview.net/forum?id=DjzBCrMBJ\_p}.

\bibitem[Liu et~al.(2022)Liu, Wang, Liu, Lasenby, Guo, and Tang]{DBLP:conf/iclr/LiuWLLGT22}
Shengchao Liu, Hanchen Wang, Weiyang Liu, Joan Lasenby, Hongyu Guo, and Jian Tang.
\newblock Pre-training molecular graph representation with 3d geometry.
\newblock In \emph{The Tenth International Conference on Learning Representations, {ICLR} 2022, Virtual Event, April 25-29, 2022}. OpenReview.net, 2022.
\newblock URL \url{https://openreview.net/forum?id=xQUe1pOKPam}.

\bibitem[Mairal et~al.(2014)Mairal, Koniusz, Harchaoui, and Schmid]{DBLP:conf/nips/MairalKHS14}
Julien Mairal, Piotr Koniusz, Za{\"{\i}}d Harchaoui, and Cordelia Schmid.
\newblock Convolutional kernel networks.
\newblock In Zoubin Ghahramani, Max Welling, Corinna Cortes, Neil~D. Lawrence, and Kilian~Q. Weinberger, editors, \emph{Advances in Neural Information Processing Systems 27: Annual Conference on Neural Information Processing Systems 2014, December 8-13 2014, Montreal, Quebec, Canada}, pages 2627--2635, 2014.
\newblock URL \url{https://proceedings.neurips.cc/paper/2014/hash/81ca0262c82e712e50c580c032d99b60-Abstract.html}.

\bibitem[M{\'{e}}moli(2011)]{DBLP:journals/focm/Memoli11}
Facundo M{\'{e}}moli.
\newblock Gromov-wasserstein distances and the metric approach to object matching.
\newblock \emph{Found. Comput. Math.}, 11\penalty0 (4):\penalty0 417--487, 2011.
\newblock \doi{10.1007/S10208-011-9093-5}.
\newblock URL \url{https://doi.org/10.1007/s10208-011-9093-5}.

\bibitem[Mohri(2018)]{mohri2018foundations}
Mehryar Mohri.
\newblock Foundations of machine learning, 2018.

\bibitem[Morris et~al.(2020)Morris, Kriege, Bause, Kersting, Mutzel, and Neumann]{DBLP:journals/corr/abs-2007-08663}
Christopher Morris, Nils~M. Kriege, Franka Bause, Kristian Kersting, Petra Mutzel, and Marion Neumann.
\newblock Tudataset: {A} collection of benchmark datasets for learning with graphs.
\newblock \emph{CoRR}, abs/2007.08663, 2020.
\newblock URL \url{https://arxiv.org/abs/2007.08663}.

\bibitem[Nikolentzos and Vazirgiannis(2020)]{DBLP:conf/nips/NikolentzosV20}
Giannis Nikolentzos and Michalis Vazirgiannis.
\newblock Random walk graph neural networks.
\newblock In Hugo Larochelle, Marc'Aurelio Ranzato, Raia Hadsell, Maria{-}Florina Balcan, and Hsuan{-}Tien Lin, editors, \emph{Advances in Neural Information Processing Systems 33: Annual Conference on Neural Information Processing Systems 2020, NeurIPS 2020, December 6-12, 2020, virtual}, 2020.
\newblock URL \url{https://proceedings.neurips.cc/paper/2020/hash/ba95d78a7c942571185308775a97a3a0-Abstract.html}.

\bibitem[Nikolentzos et~al.(2021)Nikolentzos, Siglidis, and Vazirgiannis]{DBLP:journals/jair/NikolentzosSV21}
Giannis Nikolentzos, Giannis Siglidis, and Michalis Vazirgiannis.
\newblock Graph kernels: {A} survey.
\newblock \emph{J. Artif. Intell. Res.}, 72:\penalty0 943--1027, 2021.
\newblock \doi{10.1613/JAIR.1.13225}.
\newblock URL \url{https://doi.org/10.1613/jair.1.13225}.

\bibitem[Shervashidze et~al.(2009)Shervashidze, Vishwanathan, Petri, Mehlhorn, and Borgwardt]{DBLP:journals/jmlr/ShervashidzeVPMB09}
Nino Shervashidze, S.~V.~N. Vishwanathan, Tobias Petri, Kurt Mehlhorn, and Karsten~M. Borgwardt.
\newblock Efficient graphlet kernels for large graph comparison.
\newblock In David A.~Van Dyk and Max Welling, editors, \emph{Proceedings of the Twelfth International Conference on Artificial Intelligence and Statistics, {AISTATS} 2009, Clearwater Beach, Florida, USA, April 16-18, 2009}, volume~5 of \emph{{JMLR} Proceedings}, pages 488--495. JMLR.org, 2009.
\newblock URL \url{http://proceedings.mlr.press/v5/shervashidze09a.html}.

\bibitem[Shervashidze et~al.(2011)Shervashidze, Schweitzer, van Leeuwen, Mehlhorn, and Borgwardt]{DBLP:journals/jmlr/ShervashidzeSLMB11}
Nino Shervashidze, Pascal Schweitzer, Erik~Jan van Leeuwen, Kurt Mehlhorn, and Karsten~M. Borgwardt.
\newblock Weisfeiler-lehman graph kernels.
\newblock \emph{J. Mach. Learn. Res.}, 12:\penalty0 2539--2561, 2011.
\newblock \doi{10.5555/1953048.2078187}.
\newblock URL \url{https://dl.acm.org/doi/10.5555/1953048.2078187}.

\bibitem[Shi et~al.(2021)Shi, Huang, Feng, Zhong, Wang, and Sun]{DBLP:conf/ijcai/ShiHFZWS21}
Yunsheng Shi, Zhengjie Huang, Shikun Feng, Hui Zhong, Wenjing Wang, and Yu~Sun.
\newblock Masked label prediction: Unified message passing model for semi-supervised classification.
\newblock In Zhi{-}Hua Zhou, editor, \emph{Proceedings of the Thirtieth International Joint Conference on Artificial Intelligence, {IJCAI} 2021, Virtual Event / Montreal, Canada, 19-27 August 2021}, pages 1548--1554. ijcai.org, 2021.
\newblock \doi{10.24963/IJCAI.2021/214}.
\newblock URL \url{https://doi.org/10.24963/ijcai.2021/214}.

\bibitem[Siglidis et~al.(2020)Siglidis, Nikolentzos, Limnios, Giatsidis, Skianis, and Vazirgiannis]{DBLP:journals/jmlr/SiglidisNLGSV20}
Giannis Siglidis, Giannis Nikolentzos, Stratis Limnios, Christos Giatsidis, Konstantinos Skianis, and Michalis Vazirgiannis.
\newblock Grakel: {A} graph kernel library in python.
\newblock \emph{J. Mach. Learn. Res.}, 21:\penalty0 54:1--54:5, 2020.
\newblock URL \url{https://www.jmlr.org/papers/v21/18-370.html}.

\bibitem[Togninalli et~al.(2019)Togninalli, Ghisu, Llinares{-}L{\'{o}}pez, Rieck, and Borgwardt]{DBLP:conf/nips/TogninalliGLRB19}
Matteo Togninalli, M.~Elisabetta Ghisu, Felipe Llinares{-}L{\'{o}}pez, Bastian Rieck, and Karsten~M. Borgwardt.
\newblock Wasserstein weisfeiler-lehman graph kernels.
\newblock In Hanna~M. Wallach, Hugo Larochelle, Alina Beygelzimer, Florence d'Alch{\'{e}}{-}Buc, Emily~B. Fox, and Roman Garnett, editors, \emph{Advances in Neural Information Processing Systems 32: Annual Conference on Neural Information Processing Systems 2019, NeurIPS 2019, December 8-14, 2019, Vancouver, BC, Canada}, pages 6436--6446, 2019.
\newblock URL \url{https://proceedings.neurips.cc/paper/2019/hash/73fed7fd472e502d8908794430511f4d-Abstract.html}.

\bibitem[Vayer et~al.(2019)Vayer, Courty, Tavenard, Chapel, and Flamary]{DBLP:conf/icml/VayerCTCF19}
Titouan Vayer, Nicolas Courty, Romain Tavenard, Laetitia Chapel, and R{\'{e}}mi Flamary.
\newblock Optimal transport for structured data with application on graphs.
\newblock In Kamalika Chaudhuri and Ruslan Salakhutdinov, editors, \emph{Proceedings of the 36th International Conference on Machine Learning, {ICML} 2019, 9-15 June 2019, Long Beach, California, {USA}}, volume~97 of \emph{Proceedings of Machine Learning Research}, pages 6275--6284. {PMLR}, 2019.
\newblock URL \url{http://proceedings.mlr.press/v97/titouan19a.html}.

\bibitem[Velickovic et~al.(2018)Velickovic, Cucurull, Casanova, Romero, Li{\`{o}}, and Bengio]{DBLP:conf/iclr/VelickovicCCRLB18}
Petar Velickovic, Guillem Cucurull, Arantxa Casanova, Adriana Romero, Pietro Li{\`{o}}, and Yoshua Bengio.
\newblock Graph attention networks.
\newblock In \emph{6th International Conference on Learning Representations, {ICLR} 2018, Vancouver, BC, Canada, April 30 - May 3, 2018, Conference Track Proceedings}. OpenReview.net, 2018.
\newblock URL \url{https://openreview.net/forum?id=rJXMpikCZ}.

\bibitem[Villani and Society(2003)]{villani2003topics}
C.~Villani and American~Mathematical Society.
\newblock \emph{Topics in Optimal Transportation}.
\newblock Graduate studies in mathematics. American Mathematical Society, 2003.
\newblock ISBN 9781470418045.
\newblock URL \url{https://books.google.com/books?id=MyPjjgEACAAJ}.

\bibitem[Wang et~al.(2024)Wang, Mao, Yan, Yang, Sun, Choi, Veeramani, Hu, Bowen, Cody, et~al.]{wangevolunet}
Haohui Wang, Yuzhen Mao, Yujun Yan, Yaoqing Yang, Jianhui Sun, Kevin Choi, Balaji Veeramani, Alison Hu, Edward Bowen, Tyler Cody, et~al.
\newblock Evolunet: Advancing dynamic non-iid transfer learning on graphs.
\newblock In \emph{Forty-first International Conference on Machine Learning}, 2024.

\bibitem[Weisfeiler and Lehman(1968)]{weisfeiler1968reduction}
Boris Weisfeiler and AA~Lehman.
\newblock A reduction of a graph to a canonical form and an algebra arising during this reduction.
\newblock \emph{Nauchno-Technicheskaya Informatsia}, 2\penalty0 (9):\penalty0 12--16, 1968.

\bibitem[Xinyi and Chen(2019)]{DBLP:conf/iclr/XinyiC19}
Zhang Xinyi and Lihui Chen.
\newblock Capsule graph neural network.
\newblock In \emph{7th International Conference on Learning Representations, {ICLR} 2019, New Orleans, LA, USA, May 6-9, 2019}. OpenReview.net, 2019.
\newblock URL \url{https://openreview.net/forum?id=Byl8BnRcYm}.

\bibitem[Xu et~al.(2019)Xu, Hu, Leskovec, and Jegelka]{DBLP:conf/iclr/XuHLJ19}
Keyulu Xu, Weihua Hu, Jure Leskovec, and Stefanie Jegelka.
\newblock How powerful are graph neural networks?
\newblock In \emph{7th International Conference on Learning Representations, {ICLR} 2019, New Orleans, LA, USA, May 6-9, 2019}. OpenReview.net, 2019.
\newblock URL \url{https://openreview.net/forum?id=ryGs6iA5Km}.

\bibitem[Xu et~al.(2023)Xu, Powers, Dror, Ermon, and Leskovec]{DBLP:conf/icml/XuPDEL23}
Minkai Xu, Alexander~S. Powers, Ron~O. Dror, Stefano Ermon, and Jure Leskovec.
\newblock Geometric latent diffusion models for 3d molecule generation.
\newblock In Andreas Krause, Emma Brunskill, Kyunghyun Cho, Barbara Engelhardt, Sivan Sabato, and Jonathan Scarlett, editors, \emph{International Conference on Machine Learning, {ICML} 2023, 23-29 July 2023, Honolulu, Hawaii, {USA}}, volume 202 of \emph{Proceedings of Machine Learning Research}, pages 38592--38610. {PMLR}, 2023.
\newblock URL \url{https://proceedings.mlr.press/v202/xu23n.html}.

\bibitem[Yan et~al.(2019)Yan, Zhu, Duda, Solarz, Sripada, and Koutra]{yan2019groupinn}
Yujun Yan, Jiong Zhu, Marlena Duda, Eric Solarz, Chandra Sripada, and Danai Koutra.
\newblock Groupinn: Grouping-based interpretable neural network for classification of limited, noisy brain data.
\newblock In \emph{Proceedings of the 25th ACM SIGKDD international conference on knowledge discovery \& data mining}, pages 772--782, 2019.

\bibitem[Yanardag and Vishwanathan(2015)]{DBLP:conf/kdd/YanardagV15}
Pinar Yanardag and S.~V.~N. Vishwanathan.
\newblock Deep graph kernels.
\newblock In Longbing Cao, Chengqi Zhang, Thorsten Joachims, Geoffrey~I. Webb, Dragos~D. Margineantu, and Graham Williams, editors, \emph{Proceedings of the 21th {ACM} {SIGKDD} International Conference on Knowledge Discovery and Data Mining, Sydney, NSW, Australia, August 10-13, 2015}, pages 1365--1374. {ACM}, 2015.
\newblock \doi{10.1145/2783258.2783417}.
\newblock URL \url{https://doi.org/10.1145/2783258.2783417}.

\bibitem[Ying et~al.(2018)Ying, He, Chen, Eksombatchai, Hamilton, and Leskovec]{DBLP:conf/kdd/YingHCEHL18}
Rex Ying, Ruining He, Kaifeng Chen, Pong Eksombatchai, William~L. Hamilton, and Jure Leskovec.
\newblock Graph convolutional neural networks for web-scale recommender systems.
\newblock In Yike Guo and Faisal Farooq, editors, \emph{Proceedings of the 24th {ACM} {SIGKDD} International Conference on Knowledge Discovery {\&} Data Mining, {KDD} 2018, London, UK, August 19-23, 2018}, pages 974--983. {ACM}, 2018.
\newblock \doi{10.1145/3219819.3219890}.
\newblock URL \url{https://doi.org/10.1145/3219819.3219890}.

\bibitem[You et~al.(2020)You, Chen, Sui, Chen, Wang, and Shen]{DBLP:conf/nips/YouCSCWS20}
Yuning You, Tianlong Chen, Yongduo Sui, Ting Chen, Zhangyang Wang, and Yang Shen.
\newblock Graph contrastive learning with augmentations.
\newblock In Hugo Larochelle, Marc'Aurelio Ranzato, Raia Hadsell, Maria{-}Florina Balcan, and Hsuan{-}Tien Lin, editors, \emph{Advances in Neural Information Processing Systems 33: Annual Conference on Neural Information Processing Systems 2020, NeurIPS 2020, December 6-12, 2020, virtual}, 2020.
\newblock URL \url{https://proceedings.neurips.cc/paper/2020/hash/3fe230348e9a12c13120749e3f9fa4cd-Abstract.html}.

\bibitem[Zhang et~al.(2018)Zhang, Wang, Xiang, Huang, and Nehorai]{DBLP:conf/nips/0007WX0N18}
Zhen Zhang, Mianzhi Wang, Yijian Xiang, Yan Huang, and Arye Nehorai.
\newblock Retgk: Graph kernels based on return probabilities of random walks.
\newblock In Samy Bengio, Hanna~M. Wallach, Hugo Larochelle, Kristen Grauman, Nicol{\`{o}} Cesa{-}Bianchi, and Roman Garnett, editors, \emph{Advances in Neural Information Processing Systems 31: Annual Conference on Neural Information Processing Systems 2018, NeurIPS 2018, December 3-8, 2018, Montr{\'{e}}al, Canada}, pages 3968--3978, 2018.
\newblock URL \url{https://proceedings.neurips.cc/paper/2018/hash/7f16109f1619fd7a733daf5a84c708c1-Abstract.html}.

\bibitem[Zhu et~al.(2021)Zhu, Xu, Yu, Liu, Wu, and Wang]{Zhu:2021wh}
Yanqiao Zhu, Yichen Xu, Feng Yu, Qiang Liu, Shu Wu, and Liang Wang.
\newblock {Graph Contrastive Learning with Adaptive Augmentation}.
\newblock In \emph{Proceedings of The Web Conference 2021}, WWW '21, pages 2069--2080, New York, NY, USA, April 2021. Association for Computing Machinery.
\newblock ISBN 9781450370233.
\newblock \doi{10.1145/3442381.3449802}.
\newblock URL \url{https://doi.org/10.1145/3442381.3449802}.

\end{thebibliography}
\newpage
\appendix
\section{Proof of Theorem~\ref{theorem: main}}
\label{proof:generalization}
\subsection{Learning Correct Similarity Order with Iterative Kernels}
\label{subsection:generalizationbound}
In this section, we first prove that the iterative kernel with bounded feature mapping is capable of learning the pairwise similarity order given sufficient training data, as shown by the PAC-learning framework. We consider a dataset consisting of instances $x \in \chi $, where $\chi$ is the input space. The dataset is denoted as $\chi=\left\{x_1, x_2, \ldots, x_n\right\}$. We employ a kernel function $\mathbb{K}: \chi \times \chi \rightarrow \mathbb{R}$ with an associated feature mapping $\phi: \chi \rightarrow \mathcal{H}_{\mathbb{K}}$ 
such that $\mathbb{K}(x, y) = \langle \phi(x), \phi(y) \rangle_{\mathcal{H}_{\mathbb{K}}}$. Here, $\mathcal{H}_{\mathbb{K}}$ is the Reproducing Kernel Hilbert Space (RKHS) associated with $\mathbb{K}$, and $\langle \cdot, \cdot \rangle_{\mathcal{H}{\mathbb{K}}}$ denotes the inner product in $\mathcal{H}_{\mathbb{K}}$. We assume that the feature mappings are bounded, i.e., $\|\phi(x)\|_{\mathcal{H}_{\mathbb{K}}} \leq R, \forall x \in \chi$ for some $R>0$.

Using this setup, we focus on an arbitrary fixed data point $x_i$ and compute its similarity with other data points $x_j(j \neq i)$ using the kernel function:
\begin{equation}
    S=\mathbb{K}\left(x_i, x_j\right)=\left\langle\phi\left(x_i\right), \phi\left(x_j\right)\right\rangle_{\mathcal{H}_{\mathbb{K}}}
\end{equation}

Thus, we define the function $f$ as $f\left(x_j\right)=\left\langle w, \phi\left(x_j\right)\right\rangle_{\mathcal{H}_{\mathbb{K}}}$
where $w=\phi\left(x_i\right)$.
Since $w=\phi\left(x_i\right)$ and $\left\|\phi\left(x_i\right)\right\|_{\mathcal{H}_{\mathbb{K}}} \leq R$, it follows that $\|w\|_{\mathcal{H}_{\mathbb{K}}} \leq R$. So the function $f$ representing the similarity between $x_i$ and any other point $x_j$ can be formulated as:
\begin{equation}
    f\left(x_j\right)=\left\langle\phi\left(x_i\right), \phi\left(x_j\right)\right\rangle_{\mathcal{H}_{\mathbb{K}}}=\mathbb{K}\left(x_i, x_j\right)=S_{i j} .
\end{equation}

Since we are dealing with a ranking problem, we use the hinge loss function for pairs $\left(x_j, x_k\right)(j, k \neq i)$ :
\begin{equation}
    \ell_{\text {hinge }}\left(f, x_j, x_k\right)=\max \left(0,1-s_{j k}^*\left(f\left(x_j\right)-f\left(x_k\right)\right)\right),
\end{equation}
where $s_{j k}^*= \operatorname{sign}\left(f^*\left(x_j\right)-f^*\left(x_k\right)\right)$, and $f^*$ represents the ground-truth ranking function.

Thus, the expected loss over all pairs is given by:
\begin{equation}
    L(f)=\mathbb{E}_{\left(x_j, x_k\right)}\left[\ell_{\text {hinge }}\left(f, x_j, x_k\right)\right],
\end{equation}
and the empirical loss over the sample $S$ is given by: 
\begin{equation}
    \hat{L}(f)=\frac{2}{(n-1)(n-2)} \sum_{\substack{j<k \\ j, k \neq i}} \ell_{\text {hinge }}\left(f, x_j, x_k\right),
\end{equation}
since there are $\frac{(n-1)(n-2)}{2}$ pairs among the $n-1$ data points excluding $x_i$.
To assess how well the learned function $f$ generalizes to test data, we derive a generalization bound using Rademacher complexity. We define the function class for pairs by:
\begin{equation}
    \mathcal{F}_{\text {pair }}=\left\{f\left(x_j, x_k\right)=f\left(x_j\right)-f\left(x_k\right)\right\} .
\end{equation}
which represents the similarity difference between $x_j$ and $x_k$ given the reference graph $x_i$. 

Given that  $f(x)=\langle w, \phi(x)\rangle$ with $\|w\| \leq R$ and $\|\phi(x)\| \leq R$, we can bound $f\left(x_j, x_k\right)$ by:
\begin{equation}
    \left|f\left(x_j, x_k\right)\right|=\left|f\left(x_j\right)-f\left(x_k\right)\right| \leq\left|f\left(x_j\right)\right|+\left|f\left(x_k\right)\right| \leq 2 R^2,
    \label{equ:range}
\end{equation}
since $|f(x)|=|\langle w, \phi(x)\rangle| \leq\|w\| \cdot\|\phi(x)\| \leq R \cdot R=R^2$.

Given $f\left(x_j, x_k\right)$ takes values in the range $\left[-2 R^2, 2 R^2\right]$. Therefore, following the PAC-learning framework in~\citep{mohri2018foundations}, we derive the following generalization bound. For any $\delta > 0$, with probability at least $1 - \delta$ over a sample $S$ of size $m$ drawn, the following holds for all $f \in$ $\mathcal{F}_{\text {pair }}:$

\begin{equation}
    L(f) \leq \hat{L}(f)+2 \hat{\mathfrak{R}}_S\left(\mathcal{F}_{\text {pair }}\right)+M \sqrt{\frac{\ln (1 / \delta)}{2 m}},
    \label{equ:gapfunction}
\end{equation}

where:
\begin{itemize}
    \item $\hat{\mathfrak{R}}_m\left(\mathcal{F}_{\text {pair }}\right)$ is the empirical Rademacher complexity of $\mathcal{F}_{\text {pair }}$.
    \item $M=1+2 R^2$ is an upper bound on the hinge loss function $\ell_{\text {hinge }}$.
    \item $m=\frac{(n-1)(n-2)}{2}$ is the number of pairs.
    \item $\delta$ is the confidence level.
\end{itemize}

It is observed that as $m$ increases, the term $\sqrt{\frac{\ln (1 / \delta)}{2 m}}$ decreases at a rate proportional to $\sqrt{1 / m}$. Next, we proceed by proving that the Rademacher complexity is bounded and decreases as $m$ increases.

\begin{theorem}
    Assuming $\|\phi(x)\|_{\mathcal{H}_K} \leq R$ for all $x \in \chi$, the empirical Rademacher complexity of the function class $\mathcal{F}_{\text{pair}}$ can be bounded as:
\begin{equation}
\hat{\mathfrak{R}}_{S}(\mathcal{F}_{\text{pair}}) \leq \frac{2 R^2}{\sqrt{m}}.
\end{equation}
\label{lemma:rade}
\end{theorem}

\begin{proof}
    Based on the definition of Rademacher complexity,  for function class $\mathcal{F}_{\text {pair }}$ with a sample size $m$, the complexity $\hat{\mathfrak{R}}_{S}(\mathcal{F}_{\text{pair}})$ is defined  as:
    \begin{equation}
        \hat{\mathfrak{R}}_S\left(\mathcal{F}_{\text {pair }}\right)=\mathbb{E}_\sigma\left[\sup _{f \in \mathcal{F}_{\text {pair }}} \frac{1}{m} \sum_{i=1}^m \sigma_i f\left(x_{j_i}, x_{k_i}\right)\right],
    \end{equation}
where $\sigma=\left(\sigma_1, \ldots, \sigma_m\right)$ are independent Rademacher variables (i.e., each $\sigma_k$ takes the value +1 or -1 with equal probability), and $\left\{\left(x_{j_i}, x_{k_i}\right)\right\}_{i=1}^m$ are pairs sampled from the data.

    We can express $\hat{\mathfrak{R}}_S\left(\mathcal{F}_{\text {pair }}\right)$ as:
    \begin{equation}
        \hat{\mathfrak{R}}_S\left(\mathcal{F}_{\text {pair }}\right)=\frac{1}{m} \mathbb{E}_\sigma\left[\sup _{\|w\| \leq R}\left\langle w, \sum_{i=1}^m \sigma_i\left(\phi\left(x_{j_i}\right)-\phi\left(x_{k_i}\right)\right)\right\rangle\right] .
    \end{equation}
Applying the Cauchy-Schwarz inequality:
\begin{equation}
    \langle w, v\rangle_{\mathcal{H}_{\mathbb{K}}} \leq\|w\|_{\mathcal{H}_{\mathbb{K}}}\|v\|_{\mathcal{H}_{\mathbb{K}}} \leq R\|v\|_{\mathcal{H}_{\mathbb{K}}}
\end{equation}
we have:
\begin{equation}
   \hat{\mathfrak{R}}_S\left(\mathcal{F}_{\text {pair }}\right) \leq \frac{1}{m} R \cdot \mathbb{E}_\sigma\left[\left\|\sum_{i=1}^m \sigma_i\left(\phi\left(x_{j_i}\right)-\phi\left(x_{k_i}\right)\right)\right\|\right] .
\end{equation}
   Using Jensen's inequality in the form $\mathbb{E}[\|X\|] \leq \left(\mathbb{E}\left[\|X\|^2\right]\right)^{1 / 2}$, we obtain:
   \begin{equation}
       \hat{\mathfrak{R}}_{S}(\mathcal{F}_{\text {pair }}) \leq \frac{R}{m}\left(\mathbb{E}_\sigma\left[\left\|\sum_{k=1}^m \sigma_k\left(\phi\left(x_{j_i}\right)-\phi\left(x_{k_i}\right)\right)\right\|_{\mathcal{H}_K}^2\right]\right)^{1 / 2}
   \end{equation}
 Since $\sigma_i$ are independent Rademacher variables, we have $\mathbb{E}\sigma\left[\sigma_i^2\right] = 1$. Therefore:
   \begin{equation}
        \begin{aligned}
        \mathbb{E}_\sigma\left[\left\|\sum_{i=1}^m \sigma_i\left(\phi\left(x_{j_i}\right)-\phi\left(x_{k_i}\right)\right)\right\|^2\right] &\leq \sum_{i=1}^m \mathbb{E}_\sigma\left[\sigma_i^2\right]\left\|\phi\left(x_{j_i}\right)-\phi\left(x_{k_i}\right)\right\|^2 \\
        & = \sum_{i=1}^m \left\|\phi\left(x_{j_i}\right)-\phi\left(x_{k_i}\right)\right\|^2 .
        \end{aligned}
    \end{equation}
   
    Given that $\|\phi(y)\|_{\mathcal{H}_{\mathbb{K}}} \leq R$ , we have:
    \begin{equation}
        \left\|\phi\left(x_{j_i}\right)-\phi\left(x_{k_i}\right)\right\|^2 \leq(2 R)^2=4 R^2 .
    \end{equation}

   Therefore,

\begin{equation}
    \mathbb{E}_\sigma\left[\left\|\sum_{i=1}^m \sigma_i\left(\phi\left(x_{j_i}\right)-\phi\left(x_{k_i}\right)\right)\right\|\right] \leq \sqrt{m \cdot 4 R^2}=2 R \sqrt{m} .
\end{equation}
Thus 
\begin{equation}
    \hat{\mathfrak{R}}_S\left(\mathcal{F}_{\text {pair }}\right) \leq \frac{1}{m} R \cdot 2 R \sqrt{m}=\frac{2 R^2}{m} \sqrt{m}=\frac{2 R^2}{\sqrt{m}} .
\end{equation} 

Therefore, the empirical Rademacher complexity decreases with $m$.

$\hfill\square$ 
\end{proof}

Thus, consider Equation~\ref{equ:gapfunction}:
\begin{itemize}
    \item  The term $\sqrt{\frac{\ln (1 / \delta)}{2 m}}$ decreases proportionally to $1 / \sqrt{m}$.
    \item Empirical Rademacher complexity $ \hat{\mathfrak{R}}_S\left(\mathcal{F}_{\text {pair }}\right)\leq \frac{2 R^2}{\sqrt{m}}$ also decreases proportionally to $1 / \sqrt{m}$.
\end{itemize}

Therefore, the generalization gap between $L(f)$ and $\hat{L}(f)$ decreases as $m$ increases, demonstrating that the kernel-based ranking function is PAC-learnable.

\subsection{Linking Two Key Properties to Generalizability in Binary Classification}

Building on the setup and conclusions from the previous section, we establish that the kernel can learn the correct ranking with sufficient training data. Here, we investigate how, once the correct ranking is achieved, an iterative kernel with monotonic decrease and order consistency properties ensures a decreasing error rate for test set graphs as iterations progress in binary classification tasks.

To formalize this, we define a probability space over the whole graph space where addition and scalar multiplication are defined. Let $\Omega$ be the set of all possible data points (graphs). 
Each data point $g_i \in \Omega$ represents a possible outcome, and the event of observing $g_i$ is the singleton set $\left\{g_j\right\} \subseteq \Omega$. By considering all subsets of $\Omega$, we construct an event space $\mathcal{F}$. Assuming the existence of an underlying probability distribution over these outcomes, we obtain the probability space $(\Omega, \mathcal{F}, P)$. Using this framework, let $\mu_x$ and $\mu_y$ represent the unknown mean graph in class 1 and class 2, respectively. Let $\mathbb{K}_{gt}(\cdot,\cdot) \in [0,1]$ represent the ground-truth kernel function for pairs of graphs.

Without loss of generality, we analyze the error rate for graphs belonging to class 1; the error rate for class 2 can be derived in a similar manner. We define two random variables associated with the ground-truth kernel: $\mathbf{s}_x=\mathbb{K}_{gt}({\mu_{x}, x)}$ and $\mathbf{s}_y =\mathbb{K}_{gt}({\mu_{y}, x)}$, where $x$ is a randomly sampled instance from class 1. The terms $\mathbf{s}_x (x)$ and $\mathbf{s}_y (x)$ denote specific realizations of $\mathbf{s}_x$ and $\mathbf{s}_y$, respectively, for a given realization of the graph $x$ belonging to class 1. 
Since $\mathbf{s}_x=0$ implies that there exists graph from class 1 that is completely different from $\mu_x$, which is counterintuitive, we assume that the probability density function $p(u)$, $u\in[0,1]$, of $\mathbf{s}_x$ has the following asymptotic property on the right neighborhood of $u=0$: 
\begin{equation}\label{eq:asym}
    \lim_{u\to 0^+} \frac{p(u)}{u}=C,
\end{equation}
where constant $C$ satisfies that $0\leq C<+\infty$. Let $\mathbf{\bar{s}}^{(i)}_x(x)=\mathbb{K}^{(i)}(\bar{\mu}_x^{(i)}, x)$ and $\mathbf{\bar{s}}^{(i)}_y(x)=\mathbb{K}^{(i)}(\bar{\mu}_y^{(i)}, x)$ represent the estimated realizations of $\mathbf{s}_x$ and $\mathbf{s}_y$ at iteration $i$, respectively, for a given graph realization $x$. $\bar{\mu}_x^{(i)}$ and $\bar{\mu}_y^{(i)}$ are estimates for $\mu_x$ and $\mu_y$, respectively. We assume that $\bar{\mu}_x^{(i)}$ and $\bar{\mu}_y^{(i)}$ are obtained through operations on the training graphs, where these operations are closed in the graph space. Due to the capability of the iterative kernels, we further assume that as the iteration $i \to \infty$, the sequences $\{\bar{\mu}_x^{(i)}(\cdot, \cdot)\}_{i=1}^{+\infty}$, $\{\bar{\mu}_y^{(i)}(\cdot, \cdot)\}_{i=1}^{+\infty}$ and $\{\mathbb{K}^{(i)}(\cdot, \cdot)\}_{i=1}^{+\infty}$ converge to $\mu_x$, $\mu_y$ and $\mathbb{K}_{gt}$, respectively. We assume that the convergence rate of $\{\mathbb{K}^{(i)}(\cdot, \cdot)\}_{i=1}^{+\infty}$ is much slower than the convergence rate of $\{\bar{\mu}_x^{(i)}(\cdot, \cdot)\}_{i=1}^{+\infty}$ and $\{\bar{\mu}_y^{(i)}(\cdot, \cdot)\}_{i=1}^{+\infty}$.
Furthermore, every function in the series of function $\{\mathbb{K}^{(i)}(\cdot, \cdot)\}_{i=1}^{+\infty}$ is continuous.

Denote the training data for class 1 and class 2  as $\left\{x_1, x_2, \ldots, x_n\right\} \subset \Omega$ and $\left\{y_1, y_2, \ldots, y_m\right\} \subset \Omega$, respectively. Since the learned kernels $\mathbb{K}^{(i)}$ preserve the similarity order, we can assume, without loss of generality, the following empirical ordering of the training graphs for the kernel at the $i$-th iteration:
\begin{equation}
\mathbf{\bar{s}}^{(i)}_x({x_1)}<\mathbf{\bar{s}}^{(i)}_x({x_2)}<\cdots <\mathbf{\bar{s}}^{(i)}_x({x_n)}
\end{equation}

In this setting, for an arbitrary test graph $g$ sampled from the graph space, the decision boundary of the kernel learned during the $i$-th iteration is defined as follows:
\begin{equation}
    \begin{aligned}
    &\forall g: \, \text{if} \; \mathbf{\bar{s}}^{(i)}_x(g) > \mathbf{\bar{s}}^{(i)}_y(g) \; \text{then} \, g \, \text{is classified as class 1} \\
    &\forall g: \, \text{if} \; \mathbf{\bar{s}}^{(i)}_x(g) < \mathbf{\bar{s}}^{(i)}_y(g) \; \text{then} \, g \, \text{is classified as class 2}
\end{aligned}
\end{equation}

We assume that $\mathbf{\bar{s}}^{(i)}_x(g)$ and $\mathbf{\bar{s}}^{(i)}_y(g)$, as well as $\mathbf{{s}}_x(g)$ and $\mathbf{{s}}_y(g)$, are not equal for any of the training data points.
Next, we analyze the misclassification rate for graphs in class 1.
Let the integer $k$ satisfies: $\mathbf{\bar{s}}^{(i)}_x(x_k) < \mathbf{\bar{s}}^{(i)}_y(x_k)$, while $\mathbf{\bar{s}}^{(i)}_x(x_{k+1}) > \mathbf{\bar{s}}^{(i)}_y({x_{k+1})}$ or a given iteration $i$, which means $x_k$ is the misclassified training graph in class 1 closest to the decision boundary. If the training error is 0, we take $x_0=y_1$. We note that the integer $k$ is independent of the iteration $i$ due to the consistency principle. Next, we demonstrate this using the following lemma. 

\begin{lemma}\label{lemma:con}
    For any sufficiently large iteration number $i$, there exists a fixed index $k$ such that $\mathbf{\bar{s}}^{(i)}_x(x_k) < \mathbf{\bar{s}}^{(i)}_y(x_k)$, while $\mathbf{\bar{s}}^{(i)}_x(x_{k+1}) > \mathbf{\bar{s}}^{(i)}_y({x_{k+1})}$.
\end{lemma}
\begin{proof}
    Since the series $\{\bar{\mu}_x^{(i)}\}_{i=1}^{+\infty}$ converges to $\mu_x$, it satisfies Cauchy's convergence criterion. Specifically, $\forall\delta>0$, $\exists N_1>0$, such that when $i>N_1$,
    \begin{equation}
        |\bar{\mu}_x^{(i+1)}-\bar{\mu}_x^{(i)}|<\delta.
    \end{equation}
Since we have assumed that $\mathbb{K}^{(i)}(\cdot,\cdot)$ are continuous: $\forall \epsilon>0$, $\exists \delta>0$, $\exists N_2>0$, such that for any $x$ in class 1, $\forall i>N_2$ and $|\bar{\mu}_x^{(i+1)}-\bar{\mu}_x^{(i)}|<\delta$, we have:
\begin{equation}
    \lvert\mathbb{K}^{(i+1)}(\bar{\mu}_x^{(i+1)}, x)-\mathbb{K}^{(i+1)}(\bar{\mu}_x^{(i)}, x)\rvert<\epsilon.
    \label{eq: diff}
\end{equation}

Based on Equation~\ref{eq: diff}, we have:
\begin{equation}
\resizebox{0.90\textwidth}{!}{$
    \begin{aligned}
     &\mathbf{\bar{s}}^{(i+1)}_x(x_k) - \mathbf{\bar{s}}^{(i+1)}_y(x_k) \\
     =& \mathbf{\bar{s}}^{(i+1)}_x(x_k) - \mathbb{K}^{(i+1)}(\bar{\mu}_x^{(i)}, x_k) + \mathbb{K}^{(i+1)}(\bar{\mu}_x^{(i)}, x_k)-\mathbf{\bar{s}}^{(i+1)}_y(x_k)+\mathbb{K}^{(i+1)}(\bar{\mu}_y^{(i)}, x_k)-\mathbb{K}^{(i+1)}(\bar{\mu}_y^{(i)}, x_k)\\
     <& \lvert\mathbf{\bar{s}}^{(i+1)}_x(x_k) - \mathbb{K}^{(i+1)}(\bar{\mu}_x^{(i)}, x_k)\rvert + \lvert\mathbf{\bar{s}}^{(i+1)}_y(x_k) - \mathbb{K}^{(i+1)}(\bar{\mu}_y^{(i)}, x_k)\rvert + (\mathbb{K}^{(i+1)}(\bar{\mu}_x^{(i)}, x_k)-\mathbb{K}^{(i+1)}(\bar{\mu}_y^{(i)}, x_k))
\end{aligned}$}
\label{eq:consist}
\end{equation}
Since the consistency principle holds, $\mathbb{K}^{(i+1)}(\bar{\mu}_x^{(i)}, x_k)-\mathbb{K}^{(i+1)}(\bar{\mu}_y^{(i)}, x_k)$ and $\mathbf{\bar{s}}^{(i)}_x(x_k) - \mathbf{\bar{s}}^{(i)}_y(x_k)$ have the same sign. If  $\mathbf{\bar{s}}^{(i)}_x(x_k) - \mathbf{\bar{s}}^{(i)}_y(x_k)<0$, then  $\mathbb{K}^{(i+1)}(\bar{\mu}_x^{(i)}, x_k)-\mathbb{K}^{(i+1)}(\bar{\mu}_y^{(i)}, x_k)<0$.   For a sufficiently large $N$, we select an $\epsilon>0$ such that $\epsilon< \min_{\{i>N\}}\frac{\lvert\mathbb{K}^{(i+1)}(\bar{\mu}_x^{(i)}, x_k)-\mathbb{K}^{(i+1)}(\bar{\mu}_y^{(i)}, x_k)\rvert}{2}$. Given our assumption that no training data point $x_k$ lies precisely on the decision boundary, such an $\epsilon$ is guaranteed to exist. Consequently, Equation~\ref{eq:consist} simplifies to:
\begin{equation}
    \mathbf{\bar{s}}^{(i+1)}_x(x_k) - \mathbf{\bar{s}}^{(i+1)}_y(x_k)<2\epsilon + \mathbb{K}^{(i+1)}(\bar{\mu}_x^{(i)}, x_k)-\mathbb{K}^{(i+1)}(\bar{\mu}_y^{(i)}, x_k)<0.
    \label{eq: sign_reserve}
\end{equation}
Thus, $\mathbf{\bar{s}}^{(i+1)}_x(x_k) - \mathbf{\bar{s}}^{(i+1)}_y(x_k)$ have the same sign as $\mathbf{\bar{s}}^{(i)}_x(x_k) - \mathbf{\bar{s}}^{(i)}_y(x_k)$. If $\mathbf{\bar{s}}^{(i)}_x(x_k) - \mathbf{\bar{s}}^{(i)}_y(x_k)>0$, the same conclusion can be derived in a similar manner. Since the number of training samples is finite, we can find a sufficiently large iteration such that Equation~\ref{eq: sign_reserve} holds for all $x_k$ after that iteration.
\end{proof}

Thus, some point between $\mathbf{\bar{s}}^{(i)}_x(x_k)$ and $\mathbf{\bar{s}}^{(i)}_x(x_{k+1})$ is the decision boundary  learned at the $i$-th iteration. 

Based on our assumption that the convergence rate of $\{\mathbb{K}^{(i)}(\cdot, \cdot)\}_{i=1}^{+\infty}$ is much slower than the convergence rate of $\{\bar{\mu}_x^{(i)}(\cdot, \cdot)\}_{i=1}^{+\infty}$ and $\{\bar{\mu}_y^{(i)}(\cdot, \cdot)\}_{i=1}^{+\infty}$, $\mathbf{\bar{s}}^{(i)}_x(x_k)$ is asymptotically monotonically decreasing.
\begin{equation}
    \mathbf{\bar{s}}^{(i)}_x(x_k)>\mathbf{\bar{s}}^{(i+1)}_x(x_k), \text{when } i \to +\infty
\end{equation}
Thus,
\begin{equation}
\mathbf{\bar{s}}^{(i)}_x(x_k) \to \mathbb{K}_{gt}\left(\mu_x, x_k\right)^{+} \ \text{when } i\to+\infty \implies \mathbf{\bar{s}}^{(i)}_x(x_k) > \mathbb{K}_{gt}\left(\mu_x, x_k\right)
\end{equation}
Thus, $\forall x \in$ class 1, the probability of misclassifying 
$x$ is given by: $\mathbb{P}\{\mathbf{\bar{s}}^{(i)}_x(x_k)<\mathbb{K}_{gt}(\mu_x, x_k)\}$. Specifically, we have:
\begin{equation}\label{eq:28}
\mathbb{P}\{\mathbf{s}_x<\mathbb{K}_{gt}(\mu_x, x_k)\} \leq \mathbb{P}\{\mathbf{s}_x\leq\mathbf{\bar{s}}^{(i)}_x(x_k)\}    
\end{equation}

In the following theorem, we show that the probability of misclassifying any graph from class 1 can be bounded using Markov's inequality.
\begin{theorem}
    If the iterative kernel decreases monotonically and preserves the similarity order, the probability of misclassifying a graph \( x \in \chi \) from class 1 has an upper bound that decreases monotonically as the number of iterations increases.
\end{theorem}
\begin{proof}
To upper bound the error rate, we begin by applying the transformation $\frac{1}{\mathbf{s}_x}-1$ to the original random variable $\mathbf{s}_x$. The expectation of this transformed variable is:
\begin{equation}
\begin{aligned}
    \mathbb{E}\left(\frac{1}{\mathbf{s}_x} - 1\right) &= \int_0^1 \left(\frac{1}{u} - 1\right) p(u) \, du \\
    &= \int_0^1 \frac{1}{u} \, p(u) \, du - 1 \\
\end{aligned}
\end{equation}
Since $u=0$ is singular point, we reformulate the improper integral as: $$\lim_{\eta \rightarrow 0^{+}} \int_\eta^1 \frac{1}{u}p(u) d u-1
$$ 
Here, since $ \lim_{u\to 0^+} \frac{p(u)}{u}=C$, the limit exists and thus the given integral converges, ensuring the existence of the expectation. With this result, we can apply Markov's inequality to the transformed random variable $\frac{1}{\mathbf{s}_x} - 1$. For any nonnegative $a$, we have:
\begin{equation}
    \mathbb{P}\left[(\frac{1}{\mathbf{s}_x}-1) \geqslant a\right] \leqslant \frac{\mathbb{E}\left(\frac{1}{\mathbf{s}_x}-1\right)}{a}
\end{equation}
Thus
\begin{equation}
\mathbb{P}\left[\mathbf{s}_x\leqslant  \frac{1}{1+a}\right] \leqslant \frac{\mathbb{E}\left(\frac{1}{\mathbf{s}_x}-1\right)}{a}
\end{equation}
Let $a = \frac{1}{\mathbf{\bar{s}^{(i)}}_x(x_k)} - 1$. Referring to  Equation~\ref{eq:28}, we have:
\begin{equation}
    \mathbb{P}\{\mathbf{s}_x<\mathbb{K}_{gt}(\mu_x, x_k)\} \leqslant \mathbb{P}\{\mathbf{s}_x\leq\mathbf{\bar{s}}^{(i)}_x(x_k)\}\leqslant \frac{\mathbb{E}\left(\frac{1}{\mathbf{s}_x}-1\right)}{\frac{1}{\mathbf{\bar{s}^{(i)}}_x(x_k)}-1}    
\end{equation}

Since $\mathbf{\bar{s}^{(i)}}_x(x_k)$ decreases monotonically  and $\mathbb{E}\left(\frac{1}{\mathbf{s}_x}-1\right)$ is a constant, it follows that $\mathbb{P}\{\mathbf{s}_x<\mathbb{K}_{gt}(\mu_x, x_k)\}$ has a monotonically decreasing upper bound.
\end{proof}

\section{Theoretical Verification with WL-based Kernels}

\subsection{Proof of Theorem~\ref{thm:WL}  }\label{proof:WL}
\vspace{2mm}

Following Equation~\ref{equa:wlsubtree}, the normalized WL-subtree kernel after $h$ iterations can be expressed as:

$$
\tilde{\mathbb{K}}^{(h)}_{wl\_subtree}\left(\mathcal{G}, \mathcal{G}^{\prime}\right)=\frac{\sum_{i=1}^h \left\langle\phi(\psi^{i}(\mathcal{G})), \phi(\psi^{i}\left(\mathcal{G}^{\prime})\right)\right\rangle}{\sqrt{\sum_{i=1}^h \left\langle\phi(\psi^{i}(\mathcal{G})), \phi(\psi^{i}\left(\mathcal{G})\right)\right\rangle}\sqrt{\sum_{i=1}^h \left\langle\phi(\psi^{i}(\mathcal{G}^{\prime})), \phi(\psi^{i}\left(\mathcal{G}^{\prime})\right)\right\rangle}}
$$

Next, we show by counterexample that the WL-subtree kernel may not preserve relational structure.
 
For example, consider two graphs, $\mathcal{G}$ and $\mathcal{G}^{\prime}$, in the $h$-th iteration. The color (label) set of the nodes in these graphs is denoted as $\left\{C_a, C_b, C_c, \ldots\right\}$. Graph $\mathcal{G}$ contains 200 nodes labeled $C_a$ and 4 nodes labeled $C_b$, whereas graph $\mathcal{G}^{\prime}$ contains 4 nodes labeled $C_a$ and 200 nodes labeled $C_b$. The similarity is computed using the vectors $\phi\left(\psi^i(\mathcal{G})\right)=[200,4]$ and $\phi\left(\psi^i\left(\mathcal{G}^{\prime}\right)\right)=[4,200]$. Consequently, the resulting similarity $\tilde{\mathbb{K}}^{(h)}_{wl\_subtree}$ is 0.0400.

In the next iteration, for graph $\mathcal{G}$, assume that half of the nodes currently labeled $C_a$ are relabeled to $C_c$, while the other half are relabeled to $C_d$. Additionally, all nodes labeled $C_b$ are relabeled to $C_e$. As a result, the updated histogram vector for $\mathcal{G}$ becomes $\phi\left(\psi^{i+1}(\mathcal{G})\right) = [100, 100, 4]$. Similarly, in graph $\mathcal{G}^{\prime}$, half of the nodes labeled $C_a$ are relabeled to $C_c$ and the remaining half to $C_d$, with all nodes labeled $C_b$ relabeled to $C_e$, resulting in a histogram vector of $\phi\left(\psi^{i+1}(\mathcal{G}^{\prime})\right) = [2, 2, 200]$. Here, the normalized similarity $\tilde{\mathbb{K}}^{(h)}_{wl\_subtree}$ is 0.0404, which exceeds $\tilde{\mathbb{K}}^{(h)}_{wl\_subtree}$.

This situation violates the principle of monotonic decrease, indicating that the WL-subtree kernel does not exhibit monotonicity—a phenomenon frequently observed in real-world datasets. $\hfill\square$

\subsection{Proof of Theorem~\ref{thm:WLOA} }\label{proof:WLOA}

\subsubsection{Monotonic Decrease in WLOA Kernel}
\vspace{3mm}

The WLOA kernel after $h$ iterations can be computed as follows:

$$
\mathbb{K}^{(h)}_{WLOA}\left(\mathcal{G}, \mathcal{G}^{\prime}\right)=\sum_{i=1}^h \texttt{histmin} \left\{\phi(\psi^{i}(\mathcal{G})), \phi(\psi^{i}(\mathcal{G}^{\prime}))\right\} \cdot \omega(i)
$$
where $\omega(i)$  is a monotonically increasing function. The normalized kernel is given by:

$$
\tilde{\mathbb{K}}_{WLOA}^{(h)}(\mathcal{G}, \mathcal{G}^{\prime })=\frac{\mathbb{K}_{WLOA}^{(h)}(\mathcal{G}, \mathcal{G}^{\prime })}{\sqrt{\mathbb{K}_{WLOA}^{(h)}(\mathcal{G}, \mathcal{G})}{\sqrt{\mathbb{K}_{WLOA}^{(h)}(\mathcal{G}^{\prime }, \mathcal{G}^{\prime })}}}
$$

At any iteration $i$, the expression $\texttt{histmin}  \left\{\phi(\psi^{i}(\mathcal{G})), \phi(\psi^{i}(\mathcal{G}))\right\} \cdot \omega(i)$ equals to $ \omega(i)|\mathcal{V}|$. Consequently, for the $h$-th iteration, this implies:

$$
\mathbb{K}_{W L O A}^{(h)}(\mathcal{G}, \mathcal{G})=\sum_{i=1}^h \omega(i)|\mathcal{V}|
$$

Thus, the normalized kernel value for the graph pair $\left(\mathcal{G}, \mathcal{G}^{\prime}\right)$ can be expressed as:

$$
\tilde{\mathbb{K}}_{WLOA}^{(h)}(\mathcal{G}, \mathcal{G}^{\prime })=\frac{\mathbb{K}_{WLOA}^{(h)}(\mathcal{G}, \mathcal{G}^{\prime })}{\sqrt{|\mathcal{V}||\mathcal{V}^{\prime}|} \sum_{h}^{i=1}\omega(i)}
$$

Given this, the difference between the kernel value at the $h$-th iteration, $\tilde{\mathbb{K}}_{W L O A}^{(h)}$, and the $(h+1)$-th iteration, $\tilde{\mathbb{K}}_{WLOA}^{(h+1)}$,  can be expressed as follows:
\begin{align*}\label{equation:diff}
    &\tilde{\mathbb{K}}_{WLOA}^{(h)}(\mathcal{G}, \mathcal{G}^{\prime })-\tilde{\mathbb{K}}_{WLOA}^{(h+1)}(\mathcal{G}, \mathcal{G}^{\prime })\\
    =&\frac{\mathbb{K}_{WLOA}^{(h)}(\mathcal{G}, \mathcal{G}^{\prime })}{\sqrt{|\mathcal{V}||\mathcal{V}^{\prime}|} \sum_{i=1}^{h}\omega(i)}-\frac{\mathbb{K}_{WLOA}^{(h+1)}(\mathcal{G}, \mathcal{G}^{\prime })}{\sqrt{|\mathcal{V}||\mathcal{V}^{\prime}|}\sum_{i=1}^{h+1}\omega(i)}\\
    =&\frac{1}{\sqrt{|\mathcal{V}||\mathcal{V}^{\prime}|}}\cdot(\frac{\mathbb{K}_{WLOA}^{(h)}(\mathcal{G}, \mathcal{G}^{\prime })}{\sum_{i=1}^{h}\omega(i)}-\frac{\mathbb{K}_{WLOA}^{(h)}(\mathcal{G}, \mathcal{G}^{\prime })+\texttt{histmin} \{\phi(\psi^{h+1}(\mathcal{G})), \phi(\psi^{h+1}(\mathcal{G}^{\prime})) \omega(h+1) \} }
{\sum_{i=1}^{h+1}\omega(i) }  )\\
    :=&D
\end{align*}

Considering the hierarchical structure of the refinement processes, one can deduce the following inequality:
\small
$$
\texttt{histmin} \{\phi(\psi^{i}(\mathcal{G})), \phi(\psi^{i}(\mathcal{G}^{\prime}) )\} \leq \texttt{histmin} \{\phi(\psi^{i-1}(\mathcal{G})), \phi(\psi^{i-1}(\mathcal{G}^{\prime}) )\}\leq \cdots \texttt{histmin} \{\phi(\psi^{1}(\mathcal{G})), \phi(\psi^{1}(\mathcal{G}^{\prime}) )\}
$$
\normalsize
Thus, for the kernel at the $(h+1)$-th iteration, it follows that:
\small
$$
\texttt{histmin} \{\phi(\psi^{h+1}(\mathcal{G})), \phi(\psi^{h+1}(\mathcal{G}^{\prime}) )\} \leq 
\frac{\sum_{i=1}^{h} \texttt{histmin}\{\phi(\psi^{i}(\mathcal{G})), \phi(\psi^{i}(\mathcal{G}^{\prime})) \cdot \omega(i) \} }{\sum_{i=1}^{h}\omega(i)} 
= \frac{\mathbb{K}_{WLOA}^{(h)} \left(\mathcal{G}, \mathcal{G}^{\prime}\right)  }{\sum_{i=1}^{h}\omega(i)}  
$$
\normalsize

Given this inequality, we get
\begin{equation}\label{equation:diff}
    \begin{aligned}
    D &\geq \frac{1}{\sqrt{|\mathcal{V}||\mathcal{V}^{\prime}|}}\cdot(\frac{\mathbb{K}_{WLOA}^{(h)}(\mathcal{G}, \mathcal{G}^{\prime })}{\sum_{i=1}^{h}\omega(i)}-\frac{\mathbb{K}_{WLOA}^{(h)}(\mathcal{G}, \mathcal{G}^{\prime })+\frac{\omega(h+1)}{\sum_{i=1}^{h}\omega(i)} \mathbb{K}_{WLOA}^{(h)} \left(\mathcal{G}, \mathcal{G}^{\prime}\right)}{\sum_{i=1}^{h+1}\omega(i)})\\
    &= \frac{1}{\sqrt{|\mathcal{V}||\mathcal{V}^{\prime}|}}\cdot(\frac{\mathbb{K}_{WLOA}^{(h)}(\mathcal{G}, \mathcal{G}^{\prime })}{\sum_{i=1}^{h}\omega(i)}-\frac{\sum_{i=1}^{h}\omega(i)\frac{\mathbb{K}_{WLOA}^{(h)}(\mathcal{G}, \mathcal{G}^{\prime })}{\sum_{i=1}^{h}\omega(i)}+\frac{\omega(h+1)}{\sum_{i=1}^{h}\omega(i)} \mathbb{K}_{WLOA}^{(h)} \left(\mathcal{G}, \mathcal{G}^{\prime}\right)}{\sum_{i=1}^{h+1}\omega(i)})\\
    &= \frac{1}{\sqrt{|\mathcal{V}||\mathcal{V}^{\prime}|}}\cdot\frac{\mathbb{K}_{WLOA}^{(h)} \left(\mathcal{G}, \mathcal{G}^{\prime}\right)}{\sum_{i=1}^{h}\omega(i)}\cdot(1-\frac{\sum_{i=1}^{h}\omega(i)+\omega(h+1)}{\sum_{i=1}^{h}\omega(i)+\omega(h+1)})\\
    &=0
\end{aligned}
\end{equation}

Therefore, we get $D \geq 0$. The value of Equation~\ref{equation:diff} is non-negative, indicating that WLOA is monotonically decreasing.

\subsubsection{Order Consistency in WLOA Kernel}

Assume two graph pairs satisfy the following relation:

$$
\tilde{\mathbb{K}}_{WLOA}^{(h)}(\mathcal{G}, \mathcal{G}^{\prime }) \geq \tilde{\mathbb{K}}_{WLOA}^{(h)}(\mathcal{G}, \mathcal{G}^{\prime \prime})
$$

The normalized WLOA kernel for a graph pair $\left(\mathcal{G}, \mathcal{G}^{\prime}\right)$ at iteration $h+1$ can be expressed as:
$$
\tilde{\mathbb{K}}_{WLOA}^{(h+1)}\left(\mathcal{G}, \mathcal{G}^{\prime }\right)=\frac{\mathbb{K}_{WLOA}^{(h+1)}\left(\mathcal{G}, \mathcal{G}^{\prime}\right)}{\sqrt{|\mathcal{V}|\left|\mathcal{V}^{\prime}\right|}\sum_{i=1}^{h+1}\omega(i)}=\frac{\sum_{i=1}^{h}\omega(i)}{\sum_{i=1}^{h+1}\omega(i)} \cdot \frac{\mathbb{K}_{WLOA}^{(h+1)}\left(\mathcal{G}, \mathcal{G}^{\prime}\right)}{\sqrt{|\mathcal{V}|\left|\mathcal{V}^{\prime}\right|} \sum_{i=1}^{h}\omega(i)}
$$

$\mathbb{K}_{WLOA}^{(h)}$ is monotonically increasing  with the iteration number $h$, as it follows from: $\mathbb{K}_{WLOA}^{(h+1)} = \mathbb{K}_{WLOA}^{(h)}+\texttt{histmin} \left\{\phi(\psi^{h+1}(\mathcal{G})), \phi(\psi^{h+1}(\mathcal{G}^\prime))\right\} \cdot \omega(h+1)$. Thus, we have:

\begin{align*}
    \tilde{\mathbb{K}}_{WLOA}^{(h+1)}(\mathcal{G}, \mathcal{G}^{\prime })&\geq  \frac{\sum_{i=1}^{h}\omega(i)}{\sum_{i=1}^{h+1}\omega(i)} \cdot \frac{\mathbb{K}_{WLOA}^{(h)}(\mathcal{G}, \mathcal{G}^{\prime })}{\sqrt{|\mathcal{V}||V^{\prime }|} \sum_{i=1}^{h}\omega(i)} \\
    &\geq \frac{\sum_{i=1}^{h}\omega(i)}{\sum_{i=1}^{h+1}\omega(i)} \cdot \frac{\mathbb{K}_{WLOA}^{(h)}(\mathcal{G}, \mathcal{G}^{\prime \prime })}{\sqrt{|\mathcal{V}||\mathcal{V}^{\prime \prime}|} \sum_{i=1}^{h}\omega(i)}\\
    &=\frac{\sum_{i=1}^{h}\omega(i)}{\sum_{i=1}^{h+1}\omega(i)} \cdot\tilde{\mathbb{K}}_{WLOA}^{(h)}\left(\mathcal{G}, \mathcal{G}^{\prime\prime}\right)
\end{align*}

Given the monotonic decrease property of $\tilde{\mathbb{K}}_{WLOA}^{(h)}\left(\mathcal{G}, \mathcal{G}^{\prime \prime}\right)$,we have:

$$
\tilde{\mathbb{K}}_{WLOA}^{(h+1)}\left(\mathcal{G}, \mathcal{G}^{\prime \prime}\right) \leq \tilde{\mathbb{K}}_{WLOA}^{(h)}\left(\mathcal{G}, \mathcal{G}^{\prime \prime}\right)
$$
Thus, we can conclude:

\begin{equation}
    \begin{aligned}
    \tilde{\mathbb{K}}_{WLOA}^{(h+1)}(\mathcal{G}, \mathcal{G}^{\prime })&\geq \frac{\sum_{i=1}^{h}\omega(i)}{\sum_{i=1}^{h+1}\omega(i)} \tilde{\mathbb{K}}_{WLOA}^{(h+1)}\left(\mathcal{G}, \mathcal{G}^{\prime \prime}\right)\nonumber\\
    &= \left(1-\frac{\omega(h+1)}{\sum_{i=1}^{h+1}\omega(i)}\right) \tilde{\mathbb{K}}_{WLOA}^{(h+1)}\left(\mathcal{G}, \mathcal{G}^{\prime \prime}\right)\nonumber
\end{aligned}
\end{equation}
When $\omega(i)=1$, $\lim_{h \to \infty}\frac{\omega(h+1)}{\sum_{i=1}^{h+1}\omega(i)}=\lim_{h \to \infty}\frac{h+1}{(h+2)(h+1)/2}=\lim_{h \to \infty}\frac{2}{h+2}=0$. Therefore, $\tilde{\mathbb{K}}_{WLOA}^{(h+1)}(\mathcal{G}, \mathcal{G}^{\prime })\geq \tilde{\mathbb{K}}_{WLOA}^{(h+1)}\left(\mathcal{G}, \mathcal{G}^{\prime \prime}\right)$ when $h \to \infty$. $\hfill\square$

\section{Dataset}\label{app:dataset}
In this section, we provide the statistics of the datasets used, as shown in Table~\ref{table:dataset_stat}.
\begin{table}[h!]
\caption{Dataset statistics.}\label{table:dataset_stat}
\fontsize{10.0}{11.2}\selectfont
 \setlength{\tabcolsep}{1pt}
    \centering
    \begin{tabular}{llccccc}
    \toprule
    Dataset & Task Description & \# Class & \# Size&  Ave.Nodes & Ave.Edges & Node Label. \\
    \midrule
    ogbg-molhiv   & Molecular property prediction   & 2  & 41127 & 25.5 & 54.1 & + \\
    
    \midrule  
    PROTEINS  & Enzyme classification & 2 & 1113 & 39.06 & 72.82 &+ \\
     D\&D      & Enzyme classification & 2 & 1178 & 284.32 & 715.66 & + \\
    NCI1      & Molecular classification & 2 & 4110 & 29.87 & 32.30 &+\\
    NCI109      & Molecular classification & 2 & 4127 & 29.68 & 32.13 &+ \\

    IMDB-B    & Movie venue categorization & 2 & 1000 & 19.77 & 96.53 &-\\
    
    IMDB-M    & Movie venue categorization & 3 & 1500 & 13.00 & 65.94 &-\\
    
    COLLAB   & Collaboration classification & 3 & 5000 & 74.49 & 2457.78 &-\\
    COIL-RAG   & Computer Vision & 100 & 3900 & 3.01 & 3.02 &-\\

    \midrule
    Reddit-T    & Reddit Thread Classification & 2 & 203088 & 23.93 & 24.99 &-\\
    \bottomrule
    \end{tabular}

\end{table}

\section{Detailed Set-Up}\label{app:exp_setup}
\paragraph{TU Dataset}

We restrict the hyperparameters and ensure the same architecture is used for both the base and enhanced models on the same dataset for a fair comparison. Specifically, we fixed the hidden size to 32, the number of layers to 3, and used a global mean pooling layer to generate the graph representations. The Adam optimizer was used for optimization. During the training process, we tuned both the base and enhanced models with the same search ranges: batch size \{64, 128\}, dropout rate \{0, 0.3\} and learning rate \{0.0001, 0.001, 0.01\}.,The only additional parameter for the enhanced model is the regularization term, which ranges from \{0.1, 0.5, 1, 10\}.

\vspace{3mm}
\paragraph{OGB}
Considering the complexity of the OGB dataset, we slightly expand the hyperparameter search range. Initially, we train the model with consistency loss, exploring hidden sizes \{32, 64, 128, 256\}, batch sizes \{64, 128, 256\}, and the number of layers \{3, 4\}, while keeping other parameters consistent with our experiments on the TU dataset. Subsequently, we fix the hidden size and the number of layers to ensure an identical network structure. We then repeat the parameter search process and employ the optimal settings for testing on the base model. All experiments on the OGB dataset are repeated 5 times to calculate the mean and standard deviation.
\vspace{3mm}
\paragraph{Reddit-T}

Given that the Reddit-T and OGB datasets are of comparable size, we employ similar experimental settings for training models on these datasets. The architecture is kept fixed, and we search for optimal hyperparameters for both the base model and the models incorporating consistency loss. Each model is trained and evaluated over 5 independent runs, with the mean and standard deviation of the results recorded for performance comparison.

\section{Complexity\& Scalability }
\label{app:complexity}
\subsection{Time Complexity}
\label{app:timecomplexity}
We present the time complexity analysis for our proposed consistency loss. The loss computation involves the compuation of pairwise similarities of graphs in a batch, resulting in a computation complexity of $O\left(\text{batchsize} \cdot \frac{\text{batchsize}-1}{2}\right) = O(\text{batchsize}^2)$. Given that there are $\frac{\text{datasetsize}}{\text{batchsize}}$ batches in each training epoch and that the similarities are computed between consecutive layers, the total complexity is: $$
\begin{aligned}
O(\text { loss })&=O\left\{\text { batchsize}^2 \times(\text { layernum }-1) \times\frac{\text { datasetsize }}{\text { batchsize }}\right\}\\&=O(\text { datasetsize} \times \text{batchsize} \times \text{layernum }).
\end{aligned}
$$ 
This analysis shows that the time required to compute consistency loss scales linearly with dataset size, batch size, and the number of layers. It is important to note that the training time for baseline models also scales linearly with dataset size.

\begin{wrapfigure}{r}{0.48\textwidth}\vspace{-4mm}
\includegraphics[trim=10mm 2mm 16mm 15mm, clip, width=.95\linewidth]{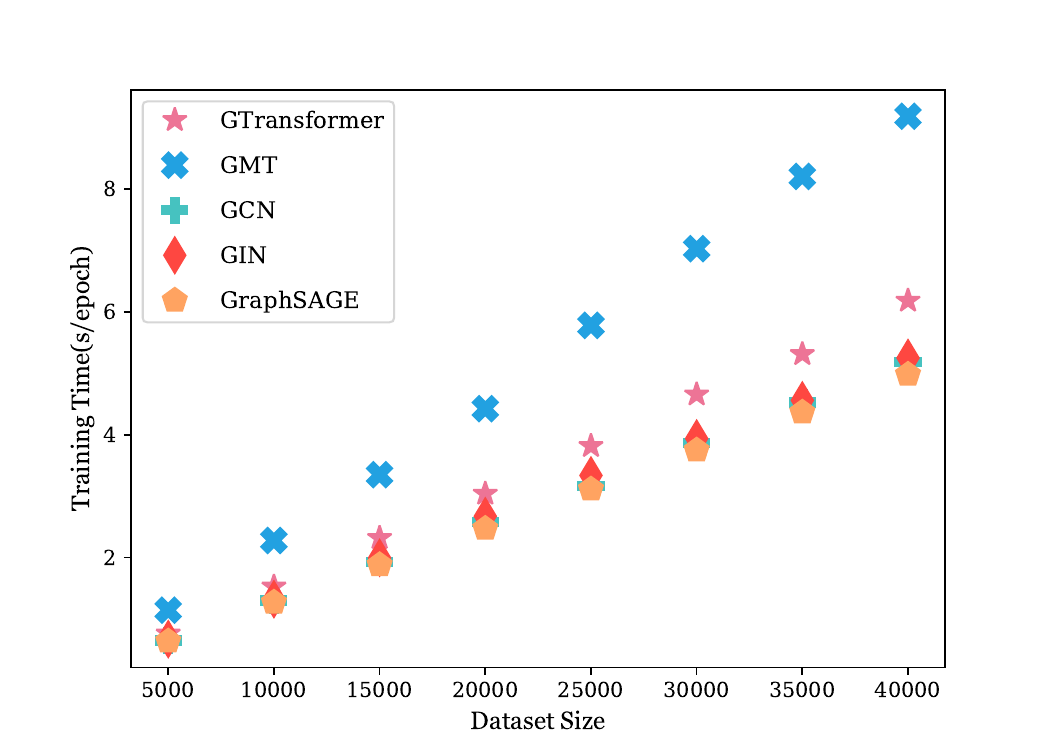}
\caption{\small{Training Cost Escalates Linearly with Dataset Size Increase}}\vspace{-4mm}
\label{pic:trainingcost}
\end{wrapfigure}

Since batch size and the number of layers are generally small compared to dataset size, our experiments primarily focus on how dataset size affects training time. We evaluate the training time of several models—GCN, GIN, GraphSAGE, GTransformer, and GMT—each enhanced with our consistency loss. This evaluation is conducted on different subsets of the ogbg-molhiv dataset, with subset sizes adjusted by varying the sampling rates. The training time, measured in seconds, are presented in Figure~\ref{pic:trainingcost} . As shown, our findings confirm that training time increases linearly with dataset size, indicating that our method maintains training efficiency comparable to baselines without adding significant time burdens.

Furthermore, we empirically measure the training time for both the baseline models and our proposed methods. Each model comprises three layers and is trained on the ogbg-molhiv dataset (40,000+ graphs) for 100 epochs. We calculate the average training time per epoch in seconds and present the results in Table~\ref{table:timecost}, showing that while the inclusion of the consistency loss slightly increases the training time, the impact is minimal.

\begin{table}[h]
\caption{ Average training time per epoch for different models on the ogbg-molhiv dataset, measured in seconds. }
\centering
\begin{tabular}{cccccc}
\toprule
     &\textbf{GMT} & \textbf{GTransformer} & \textbf{GIN} & \textbf{GCN} & \textbf{GraphSAGE} \\ 
     \midrule
\textbf{GCN}  & 8.380 & 4.937       & 4.318 & 4.221 & 3.952    \\
\textbf{GCN+$\mathcal{L}_{\text{consistency}}$ } & 8.861 & 6.358       & 5.529 & 5.382  & 5.252    \\ \bottomrule
\end{tabular}
\label{table:timecost}
\end{table}

\subsection{Space Complexity}
\label{app:spacecomplexity}
Next, we present the space complexity analysis for our consistency loss. At each iteration, the loss function requires storing two pairwise similarity matrices corresponding to two consecutive layers, which is given by:
$$
O(\text { loss })=O(\text {batchsize}^2 )
$$
Since we use stochastic gradient descent, similarity matrices are not retained for the next iteration. The consistency loss requires significantly less space than node embeddings, making the additional space requirement minimal. Table~\ref{table:memorycost} shows the peak memory usage in megabytes (MB) for different models when training on the ogbg-molhiv dataset, illustrating that the space costs are negligible.

\begin{table}[h!]

\caption{Peak memory usage for different models on the ogbg-molhiv dataset, measured in megabytes.}
\centering
\begin{tabular}{cccccc}

\toprule
&\textbf{GMT} & \textbf{GTransformer} & \textbf{GIN} & \textbf{GCN} & \textbf{GraphSAGE} \\ \midrule
\textbf{GCN} & 1334.0       & 1267.8                & 1291.3       & 1274.2       & 1288.4              \\ 
\textbf{GCN+$\mathcal{L}_{\text{consistency}}$}   & 1370.0       & 1330.6                & 1338.9       & 1320.1       & 1321.3              \\ \midrule
\textbf{\small{Cost Increase (\%)}}                     & 2.70         & 4.96                  & 3.68         & 3.60         & 2.55                \\ \bottomrule
\end{tabular}

\label{table:memorycost}
\end{table}

\subsection{Efficiency on Different Task and Structural Complexities}
\label{app:taskcomplexity}

\paragraph{Task Complexity}

We measured the runtime of the models on different subsets to evaluate how task complexity, in terms of the number of classes, influences the efficiency of the proposed method. The results are presented in Table~\ref{table:label_cost}. As demonstrated, the additional computational time remains minimal even with an increasing number of classes, suggesting that the method scales effectively with growing class complexity.

\begin{table}[h!]
\centering
\caption{\small{Average training time per epoch on REDDIT subsets with varying class complexity, measured in seconds}}
\label{table:label_cost}
\setlength{\tabcolsep}{3pt}
\begin{tabular}{ccccc}
\toprule
 & \textbf{Subset1} & \textbf{Subset2} & \textbf{Subset3} & \textbf{Fullset} \\
 & \textbf{(2 classes)} & \textbf{(3 classes)} & \textbf{(4 classes)} & \textbf{(5 classes)} \\ \midrule

\textbf{GCN} & $0.203$ & $0.345$ & $0.408$ & $0.493$  \\

\textbf{GCN+$\mathcal{L}_{\text{consistency}}$} & $0.227$ & $0.355$ & $0.430$ & $0.557$  \\

\bottomrule
\end{tabular}
\end{table}

\paragraph{Structure Complexity} 
We also conducted experiments to assess the training costs on datasets with varying structural complexities when introducing the $\mathcal{L}_{\text{consistency}}$ . The results, summarized below in Table~\ref{table:imdb_corr}, show that the additional training cost remains minimal across datasets with different structures. This demonstrates the broad applicability of the proposed method, regardless of structural complexity.

\begin{table}[h!]
\centering
\caption{Average training time per epoch for subsets of varying structural complexity from IMDB-B, measured in seconds.}
\label{table:imdb_corr}

\begin{tabular}{cccc}
\toprule

 & IMDB-B & IMDB-B & IMDB-B \\
 & \textbf{(small)} & \textbf{(medium)} & \textbf{(large)} \\ \midrule

\textbf{GCN} & $0.0308$ & $0.0311$ & $0.0321$  \\

\textbf{GCN+$\mathcal{L}_{\text{consistency}}$} & $0.0371$ & $0.0378$ & $0.0392$  \\

\bottomrule
\end{tabular}
\end{table}

\section{Efficient Consistent Learning}

\label{app:efficiency}

To further minimize the overhead of our proposed consistency loss, we examined a scenario where the consistency loss, denoted as $\mathcal{L}_{FL}$, was only applied to the first and last layers.

Building upon the experimental setup described in Section~\ref{sec:experiment}, we conducted experiments using various backbone models. The results are summarized in Table~\ref{table:efficency}. The penultimate column of this table highlights the performance gains achieved by applying the consistency loss across all layers, while the final column demonstrates the improvements observed when the consistency loss is only applied to the first and last layers.

\setlength{\tabcolsep}{3pt}
\begin{table}[h!]
\centering
\caption{Graph classification performance with improvements of $\mathcal{L}_{ALL}$ and $\mathcal{L}_{FL}$ over base models.}
\label{table:efficency}

\begin{adjustbox}{width=\textwidth,keepaspectratio}
\begin{tabular}{ccccccccc}
\toprule
& \textbf{NCI1} & \textbf{NCI109} & \textbf{PROTEINS} & \textbf{DD} & \textbf{IMDB-B} & \textbf{OGB-HIV} & $\mathcal{L}_{ALL}$ $\uparrow$ & $\mathcal{L}_{FL}$ $\uparrow$ \\ \midrule
  
\textbf{GCN+$\mathcal{L}_{FL}$} &
  75.96\tiny{$\pm 0.89$} &
  74.67\tiny{$\pm 1.11$} &
  72.97\tiny{$\pm 2.85$} &
  76.27\tiny{$\pm 1.69$} &
  74.6\tiny{$\pm 1.85$} &
  74.44\tiny{$\pm 1.42$} &
  +5.49 &
  +7.08\\
  
\textbf{GIN+$\mathcal{L}_{FL}$} &
  79.08\tiny{$\pm 1.21$} &
  77.0\tiny{$\pm 2.01$} &
  73.15\tiny{$\pm 2.76$} &
  74.07\tiny{$\pm 1.38$} &
  74.8\tiny{$\pm 4.66$} &
  74.2\tiny{$\pm 1.62$} &
  +10.95 &
  +15.12 \\
  
\textbf{GraphSAGE+$\mathcal{L}_{FL}$} &
  78.88\tiny{$\pm 2.01$} &
  74.24\tiny{$\pm 1.21$} &
  75.32\tiny{$\pm 2.46$} &
  73.90\tiny{$\pm 2.03$} &
  76.6\tiny{$\pm 1.96$} &
  80.06\tiny{$\pm 1.21$} &
  +9.10 &
  +9.71 \\
  
\textbf{GTransformer+$\mathcal{L}_{FL}$} &
  76.79\tiny{$\pm 1.24$} &
  74.38\tiny{$\pm 0.49$} &
  73.69\tiny{$\pm 2.09$} &
  75.08\tiny{$\pm 1.57$} &
  76.80\tiny{$\pm 1.60$} &
  80.53\tiny{$\pm 0.73$} &
  +9.00 &
  +8.63 \\
  
\textbf{GMT+$\mathcal{L}_{FL}$} &
  76.40\tiny{$\pm 1.00$} &
  75.64\tiny{$\pm 0.77$} &
  72.25\tiny{$\pm 3.96$} &
  73.39\tiny{$\pm 2.18$} &
  76.00\tiny{$\pm 1.36$} &
  81.05\tiny{$\pm 1.29$} &
  +6.23 &
  +5.24 \\ \bottomrule

\end{tabular}

\end{adjustbox}

\end{table}

Notably, applying the consistency loss only to the first and last layers achieves performance comparable to that of applying it across all layers, with both configurations yielding substantial improvements over the original model. This finding suggests that our proposed approach can be accelerated with minimal additional computational cost while still enhancing performance, thereby validating the effectiveness of the consistent learning principle.

\section{Similarity/Difference with Contrastive learning}

In this section, we discuss the similarities and differences between our method and graph contrastive learning.
Graph Contrastive Learning (GCL) is a self-supervised technique for graph data that emphasizes instance discrimination \citep{DBLP:conf/iclr/LinCW23,Zhu:2021wh}. A typical GCL framework generates multiple graph views via augmentations and contrasts positive samples (similar instances) with negative samples (dissimilar instances). This approach facilitates effective representation learning by capturing relationships between views, ensuring positive pairs remain close in the embedding space while distinctly separating negative pairs.

While both GCL and our method leverage graph similarity, our approach focuses on \textbf{maintaining consistency across layers, rather than solely capturing similarities as in contrastive learning.} To demonstrate this, we integrated the GraphCL technique \citep{DBLP:conf/nips/YouCSCWS20} into a GCN model (GCN+CL) and assessed its performance and layer consistency across various datasets. The results, detailed in Tables \ref{table:gcl_performance} and \ref{table:gcl_corr}, use classification accuracy and Spearman rank correlation to measure performance and consistency, respectively.

\begin{table}[h!]
\centering
\caption{Graph classification accuracy of GCN with contrastive learning applied across various datasets.}
\label{table:gcl_performance}

\begin{tabular}{cccccc}
\toprule
& \textbf{NCI1} & \textbf{NCI109} & \textbf{PROTEINS} & \textbf{D\&D} & \textbf{IMDB-B} \\ \midrule

\textbf{GCN+ CL} & 74.06 \small{$\pm 1.91$} & 73.14 \small{$\pm 1.90$} & 72.50 \small{$\pm 2.73$} & 75.80 \small{$\pm 2.09$} & 75.80 \small{$\pm 1.90$} \\

\textbf{GCN+$\mathcal{L}_{\text{consistency}}$}  & 75.12 \small{$\pm 1.19$} & 73.25 \small{$\pm 1.25$} & 75.07 \small{$\pm 5.05$} & 78.56 \small{$\pm 3.32$} & 75.85 \small{$\pm 1.82$} \\

\bottomrule
\end{tabular}
\end{table}

\begin{table}[h!]
\centering
\caption{Spearman correlation for graph representations from consecutive layers.}
\label{table:gcl_corr}

\begin{tabular}{cccccc}
\toprule
& \textbf{NCI1} & \textbf{NCI109} & \textbf{PROTEINS} & \textbf{D\&D} & \textbf{IMDB-B} \\ \midrule

\textbf{GCN+ CL} &   0.835 & 0.717 & 0.851  & 0.717 & 0.810  \\

\textbf{GCN+$\mathcal{L}_{\text{consistency}}$} &   0.859 & 0.958 &  0.946 &  0.896 &  0.907  \\

\bottomrule
\end{tabular}
\end{table}

As demonstrated by the results, our method consistently outperforms GCN+CL in both graph classification performance and in enhancing similarity consistency across layers. This underscores the significant differences between our approach and regular GCL methods.
\section{Boarder Impact }
\label{app:impact}
This paper aims to advance the field of graph learning by proposing a model-agnostic consistency learning framework. Our framework can be plugged into and improve current methods for graph classification tasks. This has potential benefits in sectors such as chemistry, bioinformatics and social  analysis, where graph classification is widely used. Additionally, we do not foresee any direct negative societal or ethical consequences stemming from our work.
\section{Limitation}
\label{app:limitation}
One limitation of our work is that the method involves additional computational costs, especially during large batch training processes. To extend our framework, sampling methodologies on data or layers can be applied during the consistency-preserving training process. By selectively sampling data points or specific layers, we can reduce the computational burden while still maintaining the effectiveness of the cross-layer consistency loss, making the framework more scalable and applicable to larger datasets.
\end{document}